\def\year{2022}\relax
\documentclass[letterpaper]{article} % DO NOT CHANGE THIS
\usepackage{aaai22}  % DO NOT CHANGE THIS
\usepackage{times}  % DO NOT CHANGE THIS
\usepackage{helvet}  % DO NOT CHANGE THIS
\usepackage{courier}  % DO NOT CHANGE THIS
\usepackage[hyphens]{url}  % DO NOT CHANGE THIS
\usepackage{graphicx} % DO NOT CHANGE THIS
\usepackage{natbib}  % DO NOT CHANGE THIS AND DO NOT ADD ANY OPTIONS TO IT
\usepackage{caption} % DO NOT CHANGE THIS AND DO NOT ADD ANY OPTIONS TO IT
\DeclareCaptionStyle{ruled}{labelfont=normalfont,labelsep=colon,strut=off} % DO NOT CHANGE THIS
\usepackage{algorithm}
\usepackage{algorithmic}

\usepackage{latexsym}
\usepackage{bm,bbm}
\usepackage{amsthm}
\usepackage{fixmath}
\usepackage{mathrsfs}
\usepackage{enumitem}
\usepackage{amsmath}
\usepackage[skip=4pt]{subcaption}
\usepackage{soul, color}
\usepackage{amsfonts}
\usepackage{booktabs}
\usepackage{multirow}
\usepackage{paralist}
\usepackage{xcolor}

\newcommand\numberthis{\addtocounter{equation}{1}\tag{\theequation}}

\newcommand{\mycolor}{\textcolor{black}}

\usepackage{newfloat}
\usepackage{listings}
\floatstyle{ruled}
\newfloat{listing}{tb}{lst}{}
\floatname{listing}{Listing}
\title{Anomaly-Injected Deep Support Vector Data Description for Text Outlier Detection}
\author{
    Zeyu You\equalcontrib\textsuperscript{\rm 1},
    Yichu Zhou\equalcontrib\textsuperscript{\rm 2},
    Tao Yang\textsuperscript{\rm 1},
    Wei Fan\textsuperscript{\rm 1}
}
\affiliations{
    Afiliations
    \textsuperscript{\rm 1}Association for Tencent America\\
    2747 Park Blvd, Palo Alto, California 94306\\
    \textsuperscript{\rm 2}Association for University of Utah\\
    201 Presidents' Cir, Salt Lake City, UT 84112\\
    {\tt youzeyu860423@gmail.com}, {\tt flyaway@cs.utah.edu}, {\tt tytaoyang@tencent.com}, {\tt davidwfan@tencent.com}
}

\begin{document}

\maketitle

\begin{abstract}
Anomaly detection or outlier detection is a common task in various domains, which has attracted significant research efforts in recent years. 
Existing works mainly focus on structured data such as numerical or categorical data; \mycolor{however}, anomaly detection on unstructured textual data is less attended.
In this work, we target the textual anomaly detection problem and propose a deep anomaly-injected support vector data description (AI-SVDD) framework. 
AI-SVDD not only learns a more compact representation of the data hypersphere but also adopts a small number of known anomalies to increase the discriminative power. 
To tackle text input, we employ a multilayer perceptron (MLP) network in conjunction with BERT to obtain enriched text representations.
We conduct experiments on three text anomaly detection applications with multiple datasets. Experimental results show that the proposed AI-SVDD is promising and outperforms existing works.

\end{abstract}

\section{Introduction}
% 1. General introduction for anomaly detection, traditional method, limited work in NLP, reasons. 
Anomaly detection \mycolor{refers} to identifying data points, events, or observations that are deviating from the normal behaviors of the datasets~\cite{ben2005outlier}. Recognizing the anomaly data is a crucial task in various fields such as identifying fraud in credit card transactions~\cite{raj2011analysis}, discriminating fraudulent claims in insurance or health care~\cite{ukil2016iot}, detecting intrusion for cyber-security~\cite{goh2017anomaly}, and monitoring rare objects or events in computer vision and surveillance video~\cite{sultani2018real}. 
Previous approaches~\cite{chandola2009anomaly} mainly focus on extracting good handcrafted features and utilizing unsupervised learning methods such as one-class support vector machine (OC-SVM)~\cite{manevitz2001one}, local outlier factor (LOF)~\cite{breunig2000lof}, and local search algorithm (LSA)~\cite{he2005optimization}. 
% Thanks to deep learning, deep anomaly detection methods~\cite{hendrycks2018deep} \hl{($\leftarrow$ two problems: 1) update bib info, accepted by ICLR-19' 2) papers cited after this are older)} are proposed and significantly improve the performance in the recent decades such as deep one-class support vector data description (OC-SVDD)~\cite{ruff2018deep}, one-class convolutional neural network (OC-CNN)~\cite{oza2018one}, anomaly detection with deep auto encoders~\cite{zhou2017anomaly}, and adversarially learned one-class classifier~\cite{sabokrou2018adversarially}.
% \hl{(maybe NLP first (rm DL approaches), then problems of anomaly detection in NLP, then discuss DL)}
% Unlike traditional anomaly detection frameworks, 

In the natural language processing (NLP) field, existing approaches mainly concentrate on recognizing specific text anomaly patterns such as hate speech~\cite{davidson2017automated,badjatiya2017deep,waseem2016hateful} or spam text~\cite{laorden2014study,miller2014twitter,savage2014anomaly}, but not on the general indistinct anomaly patterns, like domain deviation and quality variation. Detecting text anomalies is essentially hard in complex real-world applications, as the anomaly patterns are enormous, diverse, and non-trivial to define.
%%%%1. illustrate general anomaly detection senarior; 2. illustrate why classifier might not work, and problems of OC-SVDD.

To tackle general anomalies, \citet{tax2004support} proposes a support vector data description (SVDD) method, which targets to fit a spherically shaped description for the training data (from well-sampled normal class). However, without robust feature engineering, it could be difficult for SVDD to build a good data descriptor, especially for complex unstructured data. % such as natural languages. 
To this end, deep anomaly detection methods % (as reviewed in~\cite{chalapathy2019deep}) 
%\hl{($\leftarrow$ two problems: 1) update bib info, accepted by ICLR-19' 2) papers cited after this are older)} 
have been proposed and significantly improve the performance~\cite{chalapathy2019deep} such as deep one-class support vector data description (OC-SVDD)~\cite{ruff2018deep}, one-class convolutional neural network (OC-CNN)~\cite{oza2018one}, anomaly detection with deep auto encoders~\cite{zhou2017anomaly}, and adversarially learned one-class classifier~\cite{sabokrou2018adversarially}.
\mycolor{Those aforementioned approaches are designed for the the image data and cannot be directly applied to the natural language data. Here, to tackle text anomaly detection from end-to-end, we aim to build a deep network in the NLP domain and automatically discriminate the anomalies from the normal.} %, due to the difficulty of no clear definition of anomalies in natural language and the difficulty of discriminating various expression of normal natural language text from the abnormal text.
% To the best of our knowledge, there are very limited research works focusing on anomaly detection in NLP \hl{($\leftarrow$ wrong! the ideas are limited but not works)}. 
% \hl{($\leftarrow$ we need to argue how our model solve each drawback in the after paragraphs, e.g., 1) DL representations, 2) inject labels)}
% 3. application, motivation of our model, and our contributions
% \hl{(Still need to introduce SVDD first, at least SVDD = support vector data description.)}

\begin{figure}[t]
\begin{minipage}[b]{1\linewidth}
  \centering
  \centerline{\includegraphics[width=8cm]{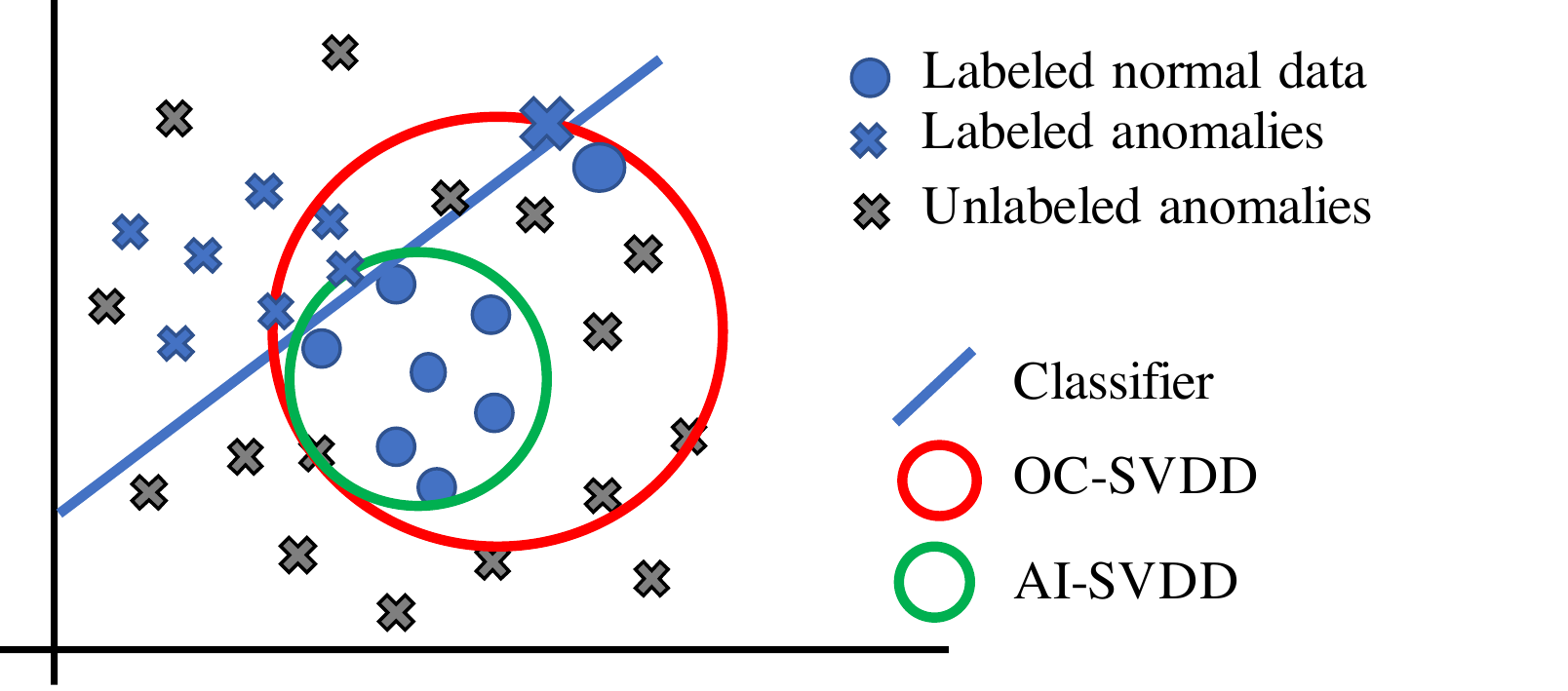}}
\end{minipage}
\caption{Motivation for AI-SVDD.}
\label{fig:model-illu}
% \vspace{-0.5cm}
\end{figure}

In this paper, motivated by the idea of deep OC-SVDD, we propose an anomaly-injected deep SVDD (AI-SVDD) framework, which provides a systematic solution for text anomaly detection. \mycolor{The motivation for AI-SVDD is illustrated in Figure~\ref{fig:model-illu}, in the 2-D feature space, anomalies might be dispersed anywhere due to enormous anomaly patterns. To detect such anomalies, a common approach is to design a model to map normal data into a compact data hypersphere, i.e., the red circle. However, without enough labeled anomalies, such a model might be sensitive to outliers or mislabelled points (upper right bigger blue circle) and make errors (many unobserved anomalies are detected as normal). Even if labeling a small portion of anomalies and treating the problem as a binary classification, a classifier can barely produce satisfactory results (see the blue line in Figure~\ref{fig:model-illu}), as it is essentially hard to cover the entire anomaly distribution. Hence, we need a robust model to differentiate normal data from diverse indistinct anomalies.} The proposed approach not only learns a more compact data description through updating the center (deep OC-SVDD uses a prefixed center), but also adopts a small number of anomaly label data to better \mycolor{discriminate} anomaly texts from the normal. More generally, the proposed AI-SVDD can be regarded as the one-class case when all labels are the same. 

The contributions are as follows: 
(i) We introduce a novel AI-SVDD deep learning framework for anomaly detection; 
(ii) We analyze the difference between deep OC-SVDD and the proposed AI-SVDD; 
(iii) We propose a systematic pipeline that combines AI-SVDD objective with deep NLP model and a solution to tackle the quadratic equality constrained optimization problem; 
(iv) We provide results and analysis on two potential application scenarios and one real-world application scenario, and we also demonstrate that the proposed deep AI-SVDD outperforms existing works.
% we explore the idea of the proposed approach \hl{($\leftarrow$ you haven't proposed anything yet)} in OC-SVDD and focus on the application of detecting the anomaly in text data containing, sentences, paragraphs or documents \hl{($\leftarrow$ each type here shall be covered in experiments, if not, rm)}. Since the anomaly text from the natural language perspective is a vague concept, a small number of human labels would be helpful and provide the guidance to better discriminates anomaly texts from the normal data. 

% \hl{(1 OC-SVDD + updated center; 2 AI-SVDD (it can be use other than NLP); 3 AI-SVDD+NLP); 4...}
% Therefore, in this work, we introduce an anomaly-injected deep SVDD (AI-SVDD) framework, which is a more general version of the deep one-class anomaly detection objective. Specifically, the AI-SVDD objective is the same as the one class objective when all labels are the same. 
% The contribution of this paper is as follows: (i) We introduce a novel AI-SVDD deep learning framework for anomaly detection in text; (ii) We provide a detailed analysis and discussion between deep OC-SVDD and our proposed AI-SVDD; (iii) We propose a systematic NLP model and the solution approach to tackle the quadratic equality constraints minimization problem; (iv) We provide two potential applications using synthetic data generation process and one real-world application results; and (v) We show how the anomaly labels can be used to improve the performance in the experiments.

% \vspace{-0.1cm}
\section{Related Work}

Anomaly detection in general is a broad topic. In this section, we review some existing text anomaly detection methods and introduce a series of deep learning based approaches that are closely related to our work.

To detect anomalies in text, \citet{manevitz2001one} propose a one-class support vector machines (OC-SVM) approach that treats document classification in a one-against-all fashion~\cite{manevitz2001one}. 
Similarly, \citet{liu2002partially} introduce a partially-supervised classification method that classifies positive class documents against all others.
For streaming text data, \citet{mahapatra2012contextual} propose a LDA-based text clustering algorithm and considers including contextual pattern as side information. 
Those traditional text anomaly detection approaches utilize feature engineering techniques such as tf-idf or ngram feature to generate text representations. 
There are also works \mycolor{that} use end-to-end models. For example,  \citet{larson2019outlier} employ sentence embedding and rank samples based on their distances to the generated mean of embeddings to detect outliers in a dialogue system; \citet{nedelchev2020treating} propose a recurrent neural network (RNN) type of encoder and decoder to model the dialogue quality and perceive dialogue evaluation as an anomaly detection task; \citet{zhuang2017identifying} and \citet{ruff2019self} focus on identifying outlier document and both utilize vector representations of words in the model.

Next, we closely review the deep support vector data description (SVDD) approaches for anomaly detection.
% deep one class svdd and semi-supervised approach
In~\citet{ruff2018deep}, an SVDD objective in combining with \mycolor{deep learning models, such as CNN} is introduced to tackle the high-dimensional, data-rich scenarios. This approach minimizes the volume of a hypersphere that encloses the network representations of the data and shows significant performance improvement of anomaly detection in MNIST~\cite{lecun1998mnist} and CIFAR-10~\cite{krizhevsky2009learning} datasets in the field of computer vision. However, unlike the original SVDD objective that jointly minimizes the hypersphere and updates the center in the feature space~\cite{tax2004support}, deep OC-SVDD considers the center of the hypersphere is known and fixed during the learning process. 
%, while the center often comes from the data that should be jointly learned in the training. In addition,
Moreover, all those aforementioned traditional and end-to-end approaches are unsupervised.

In addition, to our best knowledge, there is limited work dealing with anomaly detection in a supervised fashion. 
% In addition, to the best of our knowledge, the aforementioned anomaly detection models are in the unsupervised fashion and there is very limited work focusing on the supervised fashion in anomaly detection, which utilizes a small proportion of the labeled anomalies. 
To increase the discrimination power, a deep semi-supervised anomaly detection approach is introduced in~\citet{ruff2019deep}. It utilizes a subset of labeled data (verified by some domain expert as being normal or anomalous) together with large unlabeled data samples and shows slight performance improvement over the OC-SVDD approach. 
A clear drawback of their approach is that the learned network could project outlier data points to be enclosed in the hypersphere unexpectedly, as they assume that the majority of the unlabeled data belongs to the normal class but the quadratic objective is sensitive to outliers by natural.
% Since they assume the majority of the unlabeled data belongs to normal class, the problem they might encounter is the learned network also transform outlier data points to be enclosed also in the hypersphere. This is due to the fact that the quadratic objective is sensitive to outliers.

To this end, we propose an improved deep anomaly detection framework. It jointly minimizes the volume of a normal class data enclosed hypersphere and updates the center, and in the meanwhile, our method further adopts a small proportion of the anomaly labeled data to enhance the discrimination power.

% text anomaly detection methods

\section{Deep OC-SVDD and Revisions}
\subsection{Background and motivation}

\citet{ruff2018deep} introduces a deep one-class classification framework, which aims to train a neural network $\phi$ that minimizes the volume of a data-enclosing hypersphere centered on a predetermined projecting center ${\bf c}$. Formally, for some input space $X \subseteq \mathbb{R}^D$ and latent (feature) space $Z \subseteq \mathbb{R}^d$, let $\phi(X; {\cal W}): X\rightarrow Z$ be a $L$ hidden layers neural network with the corresponding set of weights ${\cal W} = \{{\bf W}^1
,\ldots ,{\bf W}^L\}$. The deep OC-SVDD objective can be formulated as:
\begin{equation}\label{eq:oc-svdd}
    \min_{{\cal W}} \frac{1}{n}\sum_{i=1}^n\|\phi({\bf x}_i ; {\cal W})-{\bf c}\|_2^2 + \frac{\lambda}{2}\sum_{l=1}^L\|{\bf W}^l\|_F^2,% ~\lambda > 0.
\end{equation}
where $\{ {\bf x}_i \}_1^n$ are samples on $X$, ${\bf c} \in Z$, \mycolor{$\|{\bf W}^l\|_F^2=\|{\bf W}^l\|_{2,1}^2=\sum_{a,b}|{\bf W}_{ab}^l|^2$ is the squared Euclidean norm for a matrix, and $\lambda > 0$ is the regularization hyper-parameter}.

% Intuitively, while optimizing Eq.~\eqref{eq:oc-svdd}, the center of the transformed data will also be updated. Thus, 
Rather than minimizing the volume of a data-enclosing hypersphere on a fixed center, we can further add ${\bf c}$ into the optimization and the objective becomes:
\begin{equation}\label{eq:oc-joint}
    \min_{{\cal W}, {\bf c}} \frac{1}{n}\sum_{i=1}^n\|\phi({\bf x}_i ; {\cal W})-{\bf c}\|_2^2 + \frac{\lambda}{2}\sum_{l=1}^L\|{\bf W}^l\|_F^2.
\end{equation}

In their original work, ${\bf c}$ is essentially the center of the data points through a neural network representation. That is, while jointly optimize ${\cal W}$ and ${\bf c}$ for Eq.~\eqref{eq:oc-joint}, an optimal of ${\bf c}$ can be calculated as ${\bf c}^* = \frac{1}{n}\sum_{i=1}^n \phi({\bf x}_i ; {\cal W})$. After plugging ${\bf c}^*$ back into Eq.~\eqref{eq:oc-joint}, we can further obtain the following new objective (please check Appendix
%\ref{app:simplify1} 
for more details):
\begin{align}\label{eq:oc-joint-eq}
   \nonumber \min_{{\cal W}}~ \frac{1}{2n^2}\sum_{i,j=1}^n &\| \phi({\bf x}_i ; {\cal W})- \phi({\bf x}_j ; {\cal W})\|_2^2\\
   & + \frac{\lambda}{2}\sum_{l=1}^L\|{\bf W}^l\|_F^2.
\end{align}
For preliminary exploration of Eq.~\eqref{eq:oc-joint-eq}, if we consider a vanilla network with a simple linear layer, we could finally show that the network weights would be all zero and only transform all data points onto origin. The details are shown in the supplemental material.
%\ref{supp:case_study}.

\subsection{Revision}
Instead of minimizing Eq.~\eqref{eq:oc-joint} that suffers from learning nothing, in this paper, we consider the following one-class objective:
\begin{align}\label{eq:oc-joint-alt}
    \nonumber \min_{{\cal W}, {\bf c}} & \frac{1}{n}\displaystyle{\sum_{i=1}^n} \|\phi({\bf x}_i ; {\cal W})-{\bf c}\|_2^2,\\
    & \text{s.t.~}\|{\bf W}^l\|_F^2 = 1, ~\forall l=1,2,\ldots, L.
\end{align}
Since Eq.~\eqref{eq:oc-joint-alt} minimizes the center ${\bf c}$ in addition to Eq.~\eqref{eq:oc-svdd}, the volume $\frac{1}{n}\sum_{i=1}^n\|\phi({\bf x}_i ; {\cal W})-{\bf c}^{*}\|_2^2$ would be smaller than $\frac{1}{n}\sum_{i=1}^n\|\phi({\bf x}_i ; {\cal W})-{\bf c}\|_2^2$. As a result, by introducing a joint minimization on the center ${\bf c}$, we can transform the input space to an even more compact hypersphere. 

Moreover, the one-class objective in \eqref{eq:oc-joint-alt} can be further extended to a more general scenario where the anomaly labeled data can be introduced. Therefore, we propose the AI-SVDD objective and show the differences in section~\ref{sect:method}. Note that the new objective equals to the one-class objective when all labels are the same ({\it i.e.}, all data are normal).

\section{AI-SVDD for Anomaly Detection}
\label{sect:method}
In this section, we propose to extend the aforementioned one-class objective in Eq.~\eqref{eq:oc-joint-alt} to a more general case, where the goal is to push the normal class close to a center while keeping the anomaly class far from the center. Accordingly, when additional class labels are available, anomaly detection can be more targeted and likely to be more effective and robust.

\subsection{Deep AI-SVDD objective}
In our deep AI-SVDD setting, $n$ labeled samples $\{({\bf x}_1, y_1),\ldots,({\bf x}_n, y_n)\}$ are given, where each $y_i\in \{-1, 1\}$ corresponds to anomaly label ($1$ denotes normal class, $-1$ denotes anomaly) for the sample ${\bf x}_i\in \mathbb{R}^D$. We aim to jointly find a center ${\bf c}$ and a deep neural network parameters set ${\cal W}=\{{\bf W}^1,\ldots ,{\bf W}^L\}$ with $L$ layers, such that each positive instance is close to the center and each negative instance is far from the center:
\begin{align}\label{eq:bc-joint}
    \nonumber \min_{{\cal W}, {\bf c}} & \frac{1}{n}\displaystyle{\sum_{i=1}^n} y_i \|\phi({\bf x}_i ; {\cal W})-{\bf c}\|_2^2,\\
    & \text{s.t.~} \|{\bf W}^l\|_F^2 = 1, ~\forall l=1,2,\ldots, L.
\end{align}
After solving ${\bf c}$ first, we have ${\bf c}^* = \frac{\sum_{i=1}^n y_i \phi({\bf x}_i ; {\cal W})}{\sum_{i=1}^n y_i}$. If we plug ${\bf c}^*$ back into (\ref{eq:bc-joint}) and simplify it (please check Appendix
%\ref{app:simplify2} 
for more details), we have the following new objective:
\begin{align}\label{eq:bc-joint-eq}
   \nonumber \min_{{\cal W}} & \frac{1}{2n\sum_{k=1}^n y_k}\displaystyle{\sum_{i,j=1}^n} y_i y_j \| \phi({\bf x}_i ; {\cal W})- \phi({\bf x}_j ; {\cal W})\|_2^2\\
   & \text{s.t.~} \|{\bf W}^l\|_F^2 = 1, ~\forall l=1,2,\ldots, L.
\end{align}
Note that in our proposed objective, the new center ${\bf c}^{*}$ learned by Eq.~\eqref{eq:bc-joint} takes advantage of the anomaly data in the following two folds: (i) a normal sample and an anomaly sample that are close in the embedding space cancel each other and will not contribute to the final center, which improves data robustness for the model compared with the one class case; (ii) the new center is pushed away from the anomaly points, where the simple average in the one class case can drag the center towards the anomaly when the training data is polluted with anomaly samples. Please check the supplemental material %\ref{supp:ill-centerdiff} 
for more details.

The difference between our proposed objective in Eq.~\eqref{eq:bc-joint-eq} and the deep OC-SVDD objective in Eq.~\eqref{eq:oc-svdd} is not only updating the center but also injecting the anomaly labeled data in the objective. The purpose and motivation is to introduce a more compact enclosure of the data hypersphere and in the meanwhile, increase discrimination of the anomalies from the normal data. We illustrate how different objectives perform using vanilla network on a toy synthetic data in the supplemental material.
%\ref{supp:dis-com}.

\subsection{The system and training}
\label{subsec:train}
\begin{figure}[t]
\begin{minipage}[b]{1\linewidth}
  \centering
  \centerline{\includegraphics[width=8.5cm]{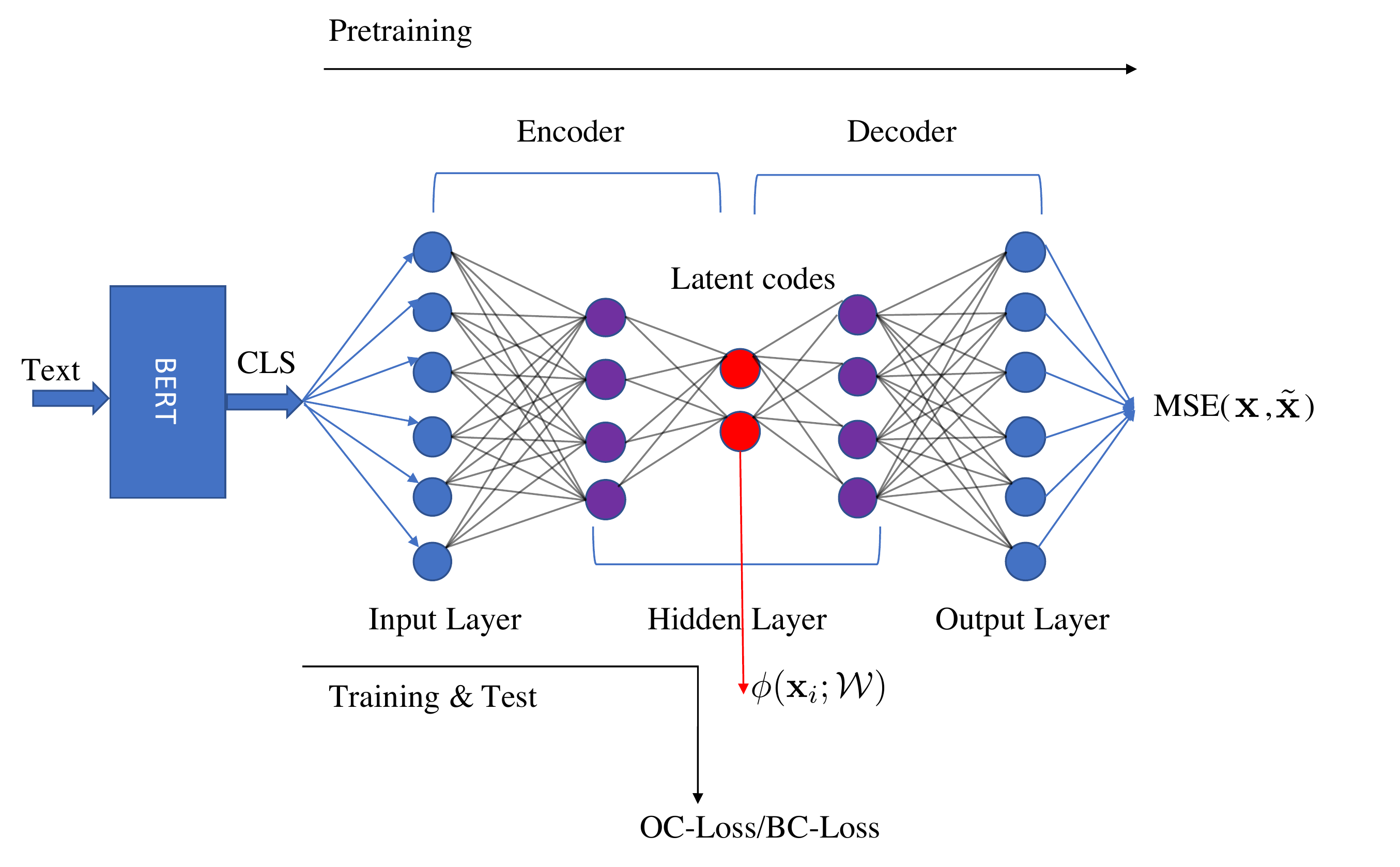}}
\end{minipage}
\caption{Diagram of the text anomaly detection system pipeline.}
\label{fig:sys-plot}
% \vspace{-0.5cm}
\end{figure}

To detect text anomalies, our target is to build the deep AI-SVDD system pipeline. We denote Eq.~\eqref{eq:oc-svdd} as the OC-Loss and the first part of Eq.~\eqref{eq:bc-joint-eq} without constraints as the BC-Loss. The goal is to optimize the deep networks corresponding to minimizing the BC-Loss. The systematic diagram of the anomaly detection is shown in Figure~\ref{fig:sys-plot}.

\noindent {\bf The system pipeline} consists of three stages: the pretraining stage, the training stage and the test stage, respectively. 
% In the pretraining stage, a multilayer perceptron (MLP) type 
In the pretraining stage, an auto-encoder is learned to transform the data from the original input space into a latent feature space. In the training stage, only the encoder part of the network is used to train on the BC-Loss or OC-Loss. In the test stage, a trained network is used to generate latent features given the test data. 

To capture a good language representation of the text, we take advantages on the pretrained bidirectional transformers (BERT) to extract initial features of the text and serve as the input data ${\bf x}_1, {\bf x}_2,\ldots, {\bf x}_n$ of the system. 
We then introduce \mycolor{auto-encoder} architecture as pretrained to learn latent encoding from the given data by minimizing the mean squared error (MSE) between the original input (BERT embeddings) and the reconstructed ones. A multilayer perceptron (MLP) network is employed in the network architecture of both encoder and decoder to transform the input into the latent space with a reduced dimension, where the hidden size, and latent dimension can be selected through cross validation~\cite{browne2000cross}.

Since our proposed BC-Loss does not involve the center ${\bf c}$, unlike the OC-Loss in~Eq.~\eqref{eq:oc-svdd}, we do not need the pretraining stage. 
We design the pretraining stage particularly for OC-Loss, which is a necessary step to obtain a good prefixed center for the deep OC-SVDD method.

\noindent{\bf Network weights minimization with quadratic equality constraints:} Since the BC-Loss contains a set of quadratic equality constraints, we rely on the projected gradient descent method~\cite{bubeck2014convex} to solve the minimization problem. For $l=1,2,\ldots, L$, the process is we first update the network parameters based on the backward error propagation step as in Eq.~\eqref{eq:bp}, and then perform a projection onto the constraints as in Eq.~\eqref{eq:proj}:
\begin{align}
    {\bf W}_{t+1}^l & = {\bf W}_t^l - \gamma \frac{\delta L({\bf W}_t^l)}{\delta {\bf W}_t^l},\label{eq:bp}\\
    {\bf W}_{t+1}^l & = \frac{{\bf W}_{t+1}^l}{\|{\bf W}_{t+1}^l\|_F},\label{eq:proj}
\end{align}
where $\gamma$ is the learning rate, ${\bf W}^l_{t}$ is the learned $l$th layer network weights at iteration $t$, ${\bf W}^l_{t+1}$ is the updated weights at the $t+1$ time step weights and $L({\bf W}_t^l)$ is the loss function evaluated at ${\bf W}^l_{t}$.
The projected gradient descent method has the similar convergence rate as that of the gradient descent method on the unconstrained objective~\cite{bubeck2014convex}.

\subsection{Inference}
In the inference stage, once the network is trained, the score for a test point ${\bf x}^{\text{test}}$ is given by the distance from $\phi( {\bf x}^{\text{test}}; {\cal W})$ to the center ${\bf c}^*=\frac{\sum_{i=1}^n y_i \phi({\bf x}_i ; {\cal W})}{\sum_{i=1}^n y_i}$ obtained from training data:
\begin{equation}\label{eq:inference}
    T({\bf x}^{\text{test}}) = \|\phi( {\bf x}^{\text{test}}; {\cal W})-{\bf c}^*\|.
\end{equation}

\section{Experiments}
\label{sect:exp}
% \vspace{-0.1cm}
In the following experiment\footnote{Code will be released upon publication.}, we consider two potential application scenarios, \mycolor{one is to detect irrelevant sentences in a topic, such as movie reviews, and the other is to detect mislabeled text data that does not belong to a known class type.} Finally, we employ a real-world medical document dataset for the application of title quality filtering.
We consider using the following baseline methods and comparing method:

\noindent {\bf The baselines} we considered here fall in the following two categories: 
\begin{inparaenum}[(i)]
\item the BERT-rank method as proposed in~\cite{larson2019outlier} that first converts sentences into embeddings, and use the training data embeddings to compute the center ${\bf c}$. To capture good representation of the text data, we use the pretrained BERT base cased model~\cite{devlin2018bert} to generate the embedding. In the test stage, we rank the test data based on the Euclidean distance of the embedding to the center from high to low for evaluation.
\item The traditional outlier detection methods such as one-class SVM (OC-SVM)~\cite{manevitz2001one}, isolation forrest (ISF)~\cite{liu2008isolation}, and local outlier factor (LOF)~\cite{breunig2000lof} use BERT embeddings as input. 
\end{inparaenum}

\noindent {\bf The comparing method} we considered is the deep OC-SVDD approach as proposed in~\citet{ruff2018deep}. Note that the work proposed in~\citet{ruff2018deep} focuses on the application domain of computer vision and can not be directly applied in text data. Here, we employ the training pipeline as described in Section~\ref{subsec:train} and learning a deep model by minimizing the loss in Eq.~\eqref{eq:oc-svdd}. 

% In terms of evaluating our proposed and the compared approaches across the experiment, we consider the following metrics:
For a comparison of text outlier detection approaches, we consider the following metrics:

\noindent {\bf Mean Average Precision (MAP)}
captures the overall precision of detecting all the anomalies, which is defined as
$\frac{1}{m}\sum_{i=1}^m\frac{|\text{anomalies above or at}  ~a_i|}{P(a_i)}$,
where $m$ is the total number of anomalies, $a_i$ is the $i$th anomaly ordered by the Euclidean distance to the center ${\bf c}^*$ in descending order, and $P(a_i)$ is the position of the $i$th anomaly in all test examples.

\noindent {\bf Recall@k} captures the percentage of retrieving the anomalies at the $k\%$ of the test data samples, which is defined as $\frac{|\text{anomalies above or at}~k\%n'|}{m}$, where $n'$ is the total number of test samples.

\noindent {\bf AUC} captures the area under the Receiver Operating Characteristic (ROC) graph of the true positive rate (TPR) against the false positive rate (FPR) by scanning through various thresholds. Here the threshold calculated by the Euclidean distance from a test sample to the center as defined in (\ref{eq:inference}).

In the topic change application, we concentrate on exploring how different levels of polluted data affects the detection performance. The BERT-rank method and the deep OC-SVDD approach are involved for comparison. We qualitatively analyze the results through ROC curves along with detailed numerical evaluations on all metrics mentioned above.
In the applications of text anomaly detection and medical text quality filtering, we compare our AI-SVDD approach with all aforementioned baselines and the comparing method quantitatively. 

% \vspace{-0.2cm}
\subsection{Text quality filtering}
% \vspace{-0.1cm}
\subsubsection{Dataset}
\mycolor{To detect whether a response is irrelevant in a specific topic text data}, we use the IMDB movie review data~\cite{maas-EtAl:2011:ACL-HLT2011} as the normal class (main topic) text data. We treat wikitext-2 data~\cite{merity2016pointer}, which contains various topics, as the anomaly text data.
For both IMDB and wikitext data, we use the Stanford NLP toolbox~\cite{qi2020stanza} to \mycolor{split} a paragraph into sentences. 
% In order to mimic the short texts in a dialog, we filter out long sentences that contains more than $50$ characters and extremely short sentences that contains less than $3$ words. 
As a result, for the normal part of the data, the processed IMDB dataset contains $9,849$ sentences for training, $991$ sentences for model development and $1,010$ sentence for testing; for anomalies, we randomly select $1,000$ sentences obtained from the wikitext data.
% An example of the generated topic change data is shown in table~\ref{tb:tc-datasets}. We can see from the example that the normal class contains various semantic meaning and rich language expression. Some of the anomaly class is quite similar to the normal class, i.e., 'The opening 5 minutes gave me hope' is normal while 'The game 's opening theme was sung by May 'n' is anomaly. 
% Unlike the anomaly detection in the vision data where the pattern of the normal class is more consistent, detecting the anomaly in text is difficult. One difficulty is to distinguish the various normal text from the anomaly text where the normal text data varies in semantic pattern.
% \begin{table}[ht]
% \small
% \centering
% \begin{tabular}{@{}cl@{}}
% \toprule
% Label & Example  \\ \midrule
%  1    & The cast is good  \\
%  1    & The opening 5 minutes gave me hope  \\
%  1    & 7 out of 10 \\
%  -1    & Vision for the PlayStation Portable       \\ 
%  -1    & The anime opening was produced by Production I.G.\\
%  -1    & The game 's opening theme was sung by May 'n\\\bottomrule
% \end{tabular}
% \caption{Examples of the data with $1$ for normal class and $-1$ for anomaly.}\label{tb:tc-datasets}
% \end{table}

% \vspace{-0.1cm}
\subsubsection{Setting}
\label{subsec:5.1setting}
To investigate how data pollution affects model performance, we consider a series of pollution proportions $p\in\{0\%, 1\%, 2\%, 4\%, 8\%\}$. More specifically, we generate the training data by combining all $9,849$ IMDB training sentences together with a total number of \mycolor{$p\times 9849$} anomaly sentences randomly selected from the wikitext data. 
The development and test datasets are generated using the $5\%$ pollution only for simplicity (also randomly selected from the rest of wikitext sentences).
We only generate each pollution proportion training dataset, development dataset and test dataset once since the normal data samples are fixed and anomaly sample size is a small proportion\footnote{Generating training data multiple times cannot cover the full anomaly distribution, and the aforementioned methods are not concentrating on learning model to fit anomaly data.}. 
In this experiment, we run our proposed deep AI-SVDD approach together with the BERT-rank method and the deep OC-SVDD approach $5$ times on different polluted training data. We report the results on the test dataset in terms of ROC, MAP, Recall@5 and AUC, which are summarized in Figure~\ref{fig:exp5.1roc} and Table~\ref{tb:exp5.1-perf}.
% To evaluate the proposed method and the aforementioned baseline and comparing methods, 
% We train each model in different $p$ polluted data and test on the $5\%$ anomaly proportion data and show the performance in terms of MAP, Recall@5, ROC an AUC.

% \vspace{-0.1cm}  
\subsubsection{Result and analysis}
\begin{figure}
  \centering
  \hspace{-0.5cm}
  \begin{subfigure}[b]{0.53\columnwidth}
 \centering
\resizebox{1 \textwidth}{!}{
  \includegraphics[width=1\textwidth]{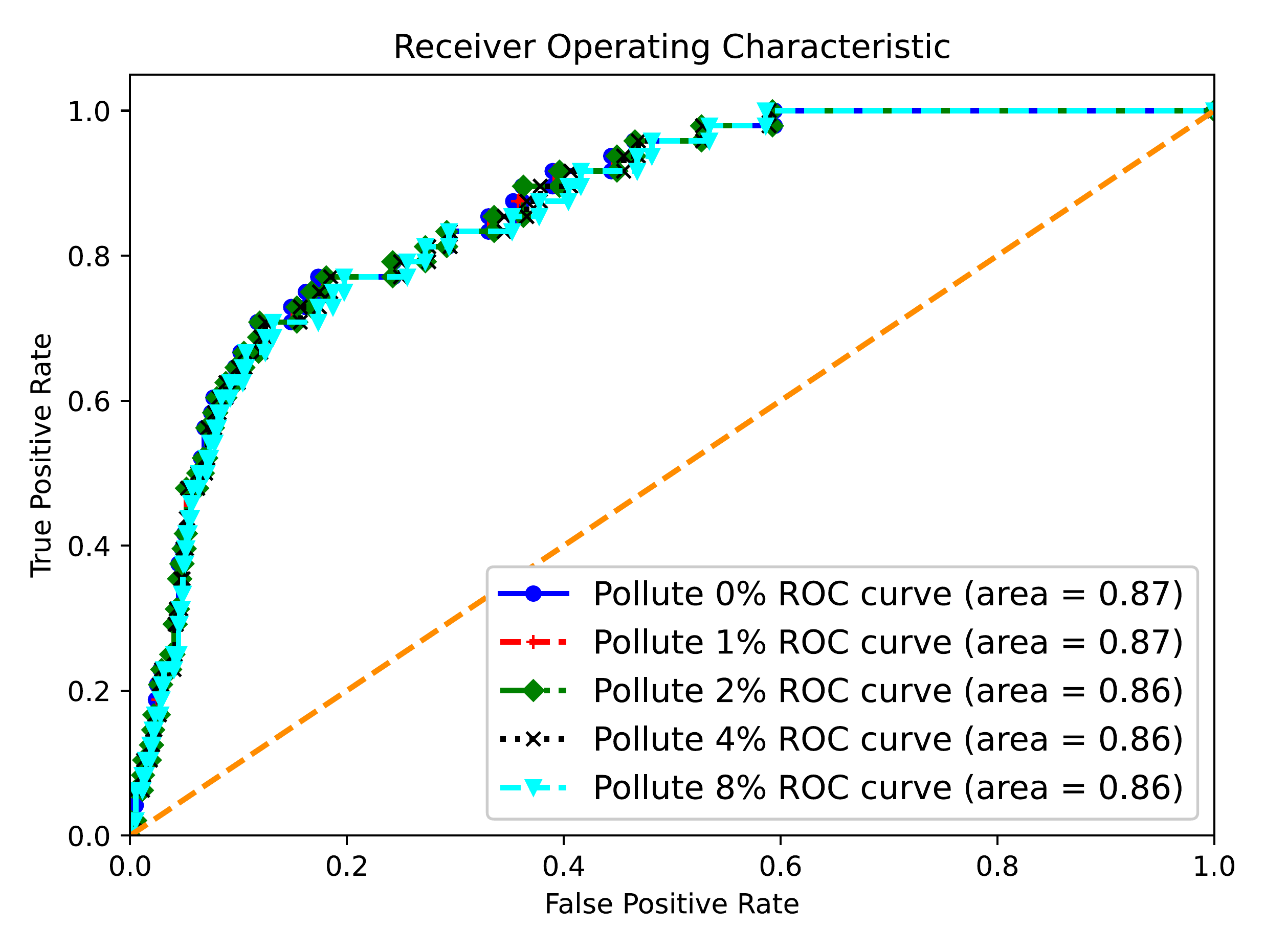}
  }
  \caption{BERT-rank ROC}
  \end{subfigure}\hspace{-0.5cm}
~
 \begin{subfigure}[b]{0.53\columnwidth}
 \centering
\resizebox{1 \textwidth}{!}{
  \includegraphics[width=1\textwidth]{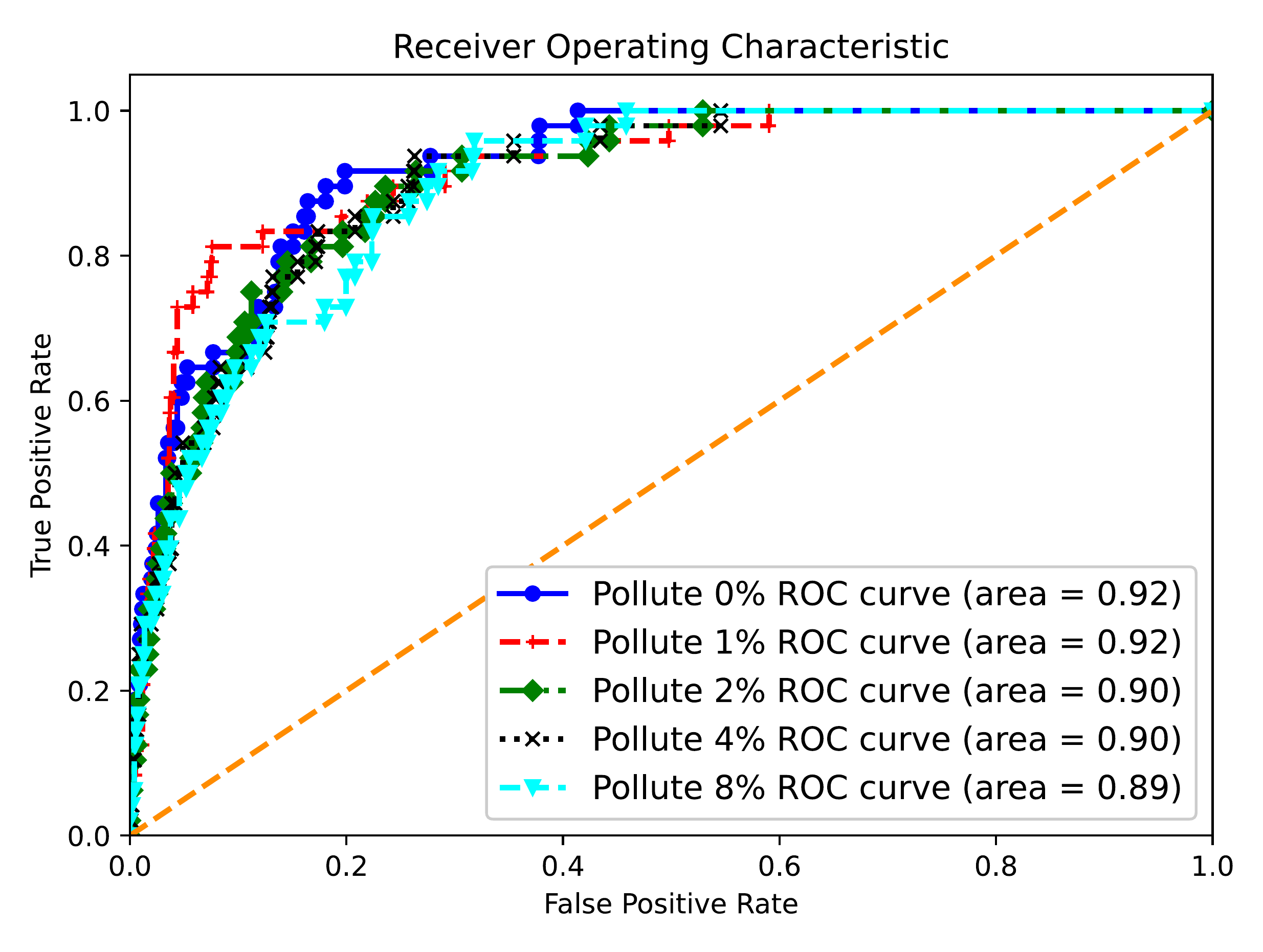}
  }
  \caption{OC-SVDD ROC}
  \end{subfigure}
  \\
  \hspace{-0.5cm}
 \begin{subfigure}[b]{0.53\columnwidth}
 \centering
\resizebox{1 \textwidth}{!}{
  \includegraphics[width=0.8\textwidth]{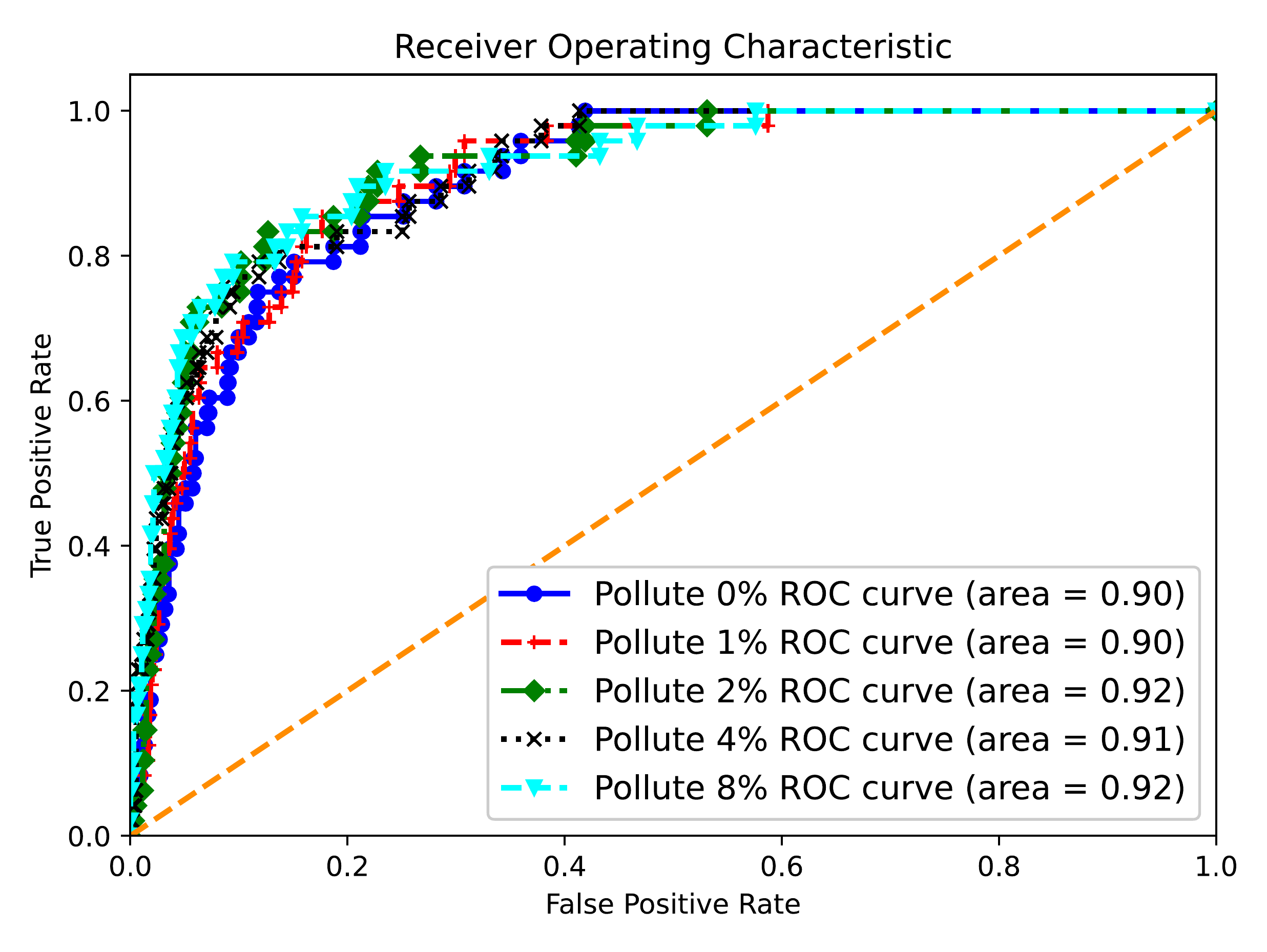}
  }
  \caption{AI-SVDD ROC}
  \end{subfigure}\hspace{-0.5cm}
  ~
 \begin{subfigure}[b]{0.53\columnwidth}
 \centering
\resizebox{1 \textwidth}{!}{
  \includegraphics[width=0.8\textwidth]{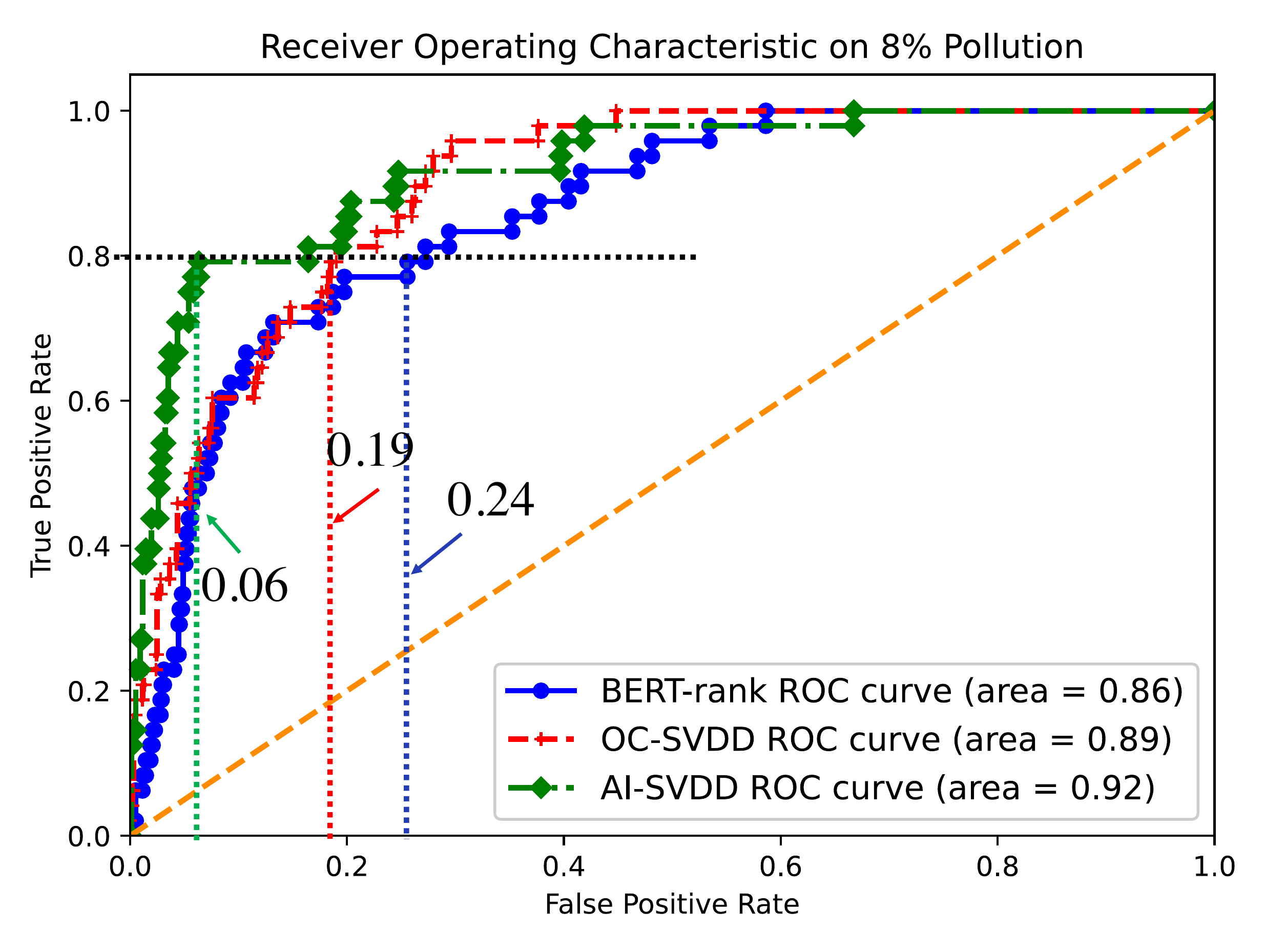}
  }
  \caption{Methods comparison}
  \end{subfigure}
  \caption{ROC curves on the test data with various $p$: (a) Baseline method(BERT-rank) (b) Deep OC-SVDD method~\cite{ruff2018deep} (c) Deep AI-SVDD method (ours) (d) Comparison of test ROC performance of all three methods on a $8\%$-polluted training data.}
  \label{fig:exp5.1roc}
%   \vspace{-0.3cm}
\end{figure}
  
Figure~\ref{fig:exp5.1roc} \mycolor{shows the ROC curves of different methods in different pollution proportion $p$ settings.}
We can observe that the AUCs of the $5$ ROC curves of the BERT-rank method does not change much across different pollution proportions, while the AUCs of the OC-SVDD method reduces gradually when data pollution proportion increases. However, the AUCs of our proposed approach increases slightly when data pollution proportion increased from $0\%$ to $8\%$, as our AI-SVDD model takes advantage of the labeled anomalies. 
Figure~\ref{fig:exp5.1roc}(d) illustrates the ROC curves of the aforementioned three methods on a $8\%$-polluted training data. 
The results illustrate that our proposed approach achieves a $3\%$ increasing in AUC compared with OC-SVDD and $6\%$ increase compared to BERT-rank method. 
In addition, by reaching $80\%$ TPR, our AI-SVDD method only introduces $6\%$ false alarms (FPR) while BERT-rank introduces $24\%$ false alarms and OC-SVDD brings in $19\%$ false alarms, respectively.

\begin{table}[t]
\setlength{\tabcolsep}{2pt}
% \vspace{-0.2cm}
\fontsize{8}{9}\selectfont
\centering
\begin{tabular}{@{}cccccc@{}}
\toprule
Method & p=0$\%$   & p=1$\%$    & p=2$\%$   & p=4$\%$ & p=8$\%$ \\\midrule
MAP \\\midrule
BERT-rank~\protect\footnotemark[3]& 24.9(0.0)  & 24.6(0.0)    & 24.6(0.0)    & 24.3(0.0)   &23.7(0.0)      \\
OC-SVDD& {\bf 46.0(2.0)}& {\bf 44.2(2.7)} & {\bf 40.8(1.8)} & {\bf 38.6(1.8)} & 34.3(2.2)   \\   
AI-SVDD& 30.0(1.3) &  33.7(2.7)   & 36.3(2.3)   & {\bf 38.2(3.0) }  & {\bf 49.1(2.8)}         \\\midrule
Recall@5\\ \midrule
BERT-rank~\protect\footnotemark[3]& 25.0(0.0)  & 25.0(0.0)    & 25.0(0.0)    & 22.9(0.0)   &22.9(0.0)        \\
OC-SVDD& {\bf 47.5(3.3)} & {\bf 47.5(3.3)}  & {\bf 46.3(2.0)}  & 43.8(3.5) & 36.6(1.6)  \\
AI-SVDD& 39.6(2.3)  & 42.1(2.8)   & 42.5(2.1)   & {\bf 46.7(4.2)}  & {\bf 50.0(2.9)}      \\\midrule 
AUC \\\midrule
BERT-rank~\protect\footnotemark[3]& 86.6(0.0)  & 86.6(0.0)    & 86.5(0.0)    & 86.3(0.0)  & 85.9(0.0)        \\
OC-SVDD& {\bf 91.2(0.4)} & {\bf 90.9(1.1)}   & 90.4(0.6)   & 89.3(0.6)  
&88.8(0.8)       \\
AI-SVDD& 89.1(0.9) &  90.0(0.6)  &{\bf 90.5(0.8)}   & {\bf 91.3(0.8)}  & {\bf 92.2(0.5)}        \\\bottomrule
\end{tabular}
\caption{Anomaly detection performance of 5 runs (mean and std.) on test data with various polluted training data (all the reported digits are in percentile $\%$).}\label{tb:exp5.1-perf}
% \vspace{-0.3cm}
\end{table}
\footnotetext[3]{Results of the BERT-rank method are consistent without variations, as it only calculates the average of embeddings of all training samples does not change across different runs.}

% \begin{table}[ht]
% \setlength{\tabcolsep}{2pt}

% % \vspace{-0.2cm}
% \fontsize{8}{9}\selectfont
% \centering
% \begin{tabular}{@{}cccccc@{}}
% \toprule
% Method & p=0$\%$   & p=1$\%$    & p=2$\%$   & p=4$\%$ & p=8$\%$ \\\midrule
% MAP \\\midrule
% BERT-rank& 38.3 & 38.0   & 37.9   & 37.4  &36.7    \\
% OC-SVDD& {\bf 63.1}& {\bf 50.8} & 45.7 & 43.7 & 39.1   \\   
% AI-SVDD& 47.1 &  49.1   & {\bf 53.9}   & {\bf 57.8}  & {\bf 74.8}         \\\midrule
% Recall@5\\ \midrule
% BERT-rank& 37.5 & 37.5   & 37.5   & 37.5  &37.5       \\
% OC-SVDD& {\bf 63.8} & {\bf 52.5}  & 48.8  & 44.6 & 39.2  \\
% AI-SVDD& 47.9  & 52.1   & {\bf 54.2}   & {\bf 57.1}  & {\bf 67.9}      \\\midrule 
% AUC \\\midrule
% BERT-rank& 94.7 & 94.6   & 94.5   & 94.4 & 94.2       \\
% OC-SVDD& {\bf 97.2} & 94.6   & 94.7   & 93.9 &92.9       \\
% AI-SVDD& 95.2 &  {\bf 95.7}  &{\bf 95.7}   & {\bf 96.3}  & {\bf 97.6}        \\\bottomrule
% \end{tabular}
% \caption{Anomaly detection performance of 5 runs (mean and std.) on test data with various polluted training data (all the reported digits are shown in percentile$\%$).}\label{tb:exp5.1-perf}
% \vspace{-0.2cm}
% \end{table}

% Table~\ref{tb:exp5.1-perf} shows the mean and standard deviation of the detection performance in terms of MAP, Recall@5 and AUC on the $5\%$ polluted test data on various methods. 
Table~\ref{tb:exp5.1-perf} shows the detection performance in terms of MAP, Recall@5 and AUC on the $5\%$-polluted test data on different methods. 
For the baseline BERT-rank method, the performance in all metrics reduce slightly when the pollution proportion in the training data is increasing. But this method in general, underperforms the anomaly detection task compared with OC-SVDD and AI-SVDD methods. %, since the baseline method only changes the center in the training. 
% when data is not polluted, OC best, but drop significantly when polluted
When data is not polluted, the deep OC-SVDD outperforms other methods but its performance reduces gradually when the pollution proportion is increasing, 
% especially when the proportions change from $0\%$ to $1\%$. This is due to the fact that its quadratic objective, which minimizes the distance between the learned embedding and the center,  is sensitive to those polluted outliers. 
on the contrary, the MAP, Recall@5 and AUC scores of our proposed deep AI-SVDD method significantly increase when the data pollution proportion is increasing, especially when the proportions change from $0\%$ to $1\%$. 
When the data is $8\%$-polluted, our deep AI-SVDD method achieves the best overall performance with average scores of $49.1\%$, $50.0\%$ and $92.2\%$, in terms of MAP, Recall@5 and AUC, respectively. 

Note that when the data is $0\%$-polluted, our approach is the same as the revised one-class objective in Eq.~\eqref{eq:oc-joint-alt}, which differs from the OC-SVDD objective on how we treat the projection center. 
% We suspect the downgrade performance of our proposed method (updating center) in $0\%$ polluted data, when compared with OC-SVDD (prefixed center), is due to two reasons: 
From Table~\ref{tb:exp5.1-perf}, we can observe a clear performance gap while using those two objectives, we suspect this is majorly due to the following two reasons: 
(i) The deep OC-SVDD involves a pretraining network and it needs to be tuned carefully for finding a good prefixed center before optimizing the OC-Loss. However, such a method is very sensitive to the choice of the center. When the center is randomly initialized, we can easily observe a significant performance decrease. 
(ii) Our AI-SVDD objective minimizes the average pair-wise squared distances of all training data in the latent space, without the anomaly label points, the model can extremely overfit the training data.

% \vspace{-0.2cm}
\subsection{Unknown class detection}

\begin{table*}[]
\centering
\tiny
\setlength{\tabcolsep}{3pt}
\renewcommand{\arraystretch}{0.5}
\begin{tabular}{@{}cc|ccccc|ccccc|ccccc@{}}
\toprule
\multirow{2}{*}{Models}       & \multirow{2}{*}{Dataset} & \multicolumn{5}{c|}{MAP}          & \multicolumn{5}{c|}{Recall@5}     & \multicolumn{5}{c}{AUC}          \\ 
                              &                          & 0    & 1    & 2    & 4    & 8    & 0    & 1    & 2    & 4    & 8    & 0    & 1    & 2    & 4    & 8    \\ \midrule
\multirow{3}{*}{OC-SVM}       & AG                       & 4.7(0.0)   & 5.0(0.0)   & 4.6(0.0)   & 4.8(0.0)   & 4.2(0.0)   & 4.2(0.0)   & 3.2(0.0)   & 5.3(0.0)   & 6.3(0.0)   & 3.2(0.0)   & 50.0(0.0)  & 52.8(0.0)  & 45.3(0.0)  & 49.2(0.0)  & 44.8(0.0)  \\
                              & Yelp                     & 5.9(0.0)   & 6.1(0.0)   & 5.2(0.0)   & 5.5(0.0)   & 5.4(0.0)   & 7.3(0.0)   & {\bf 9.1(0.0) }  & 3.7(0.0)   & 6.7(0.0)   & 7.9(0.0)   & 54.8(0.0)  & 56.0(0.0)  & 51.9(0.0)  & 52.7(0.0)  & 50.0(0.0)  \\
                              & RCT                      &   7.6(0.0)    &  7.2(0.0)     &  6.5(0.0)     &  7.4(0.0)     & 7.1(0.0)      &  10.5(0.0)     & 10.5(0.0)      & 8.3(0.0)      & 13.2(0.0)      &   11(0.0)    & 56.3(0.0)      &  57(0.0)     &  54.5(0.0)     & 55.3(0.0)       & 56(0.0)      \\ \midrule

\multirow{3}{*}{ISF}          & AG                       & 4.5(0.3)  & 4.4(0.1)  & 4.6(0.2)  & 4.2(0.2)  & 4.2(0.3)  & 1.9(1.0)  & 1.1(0.7)  & 1.1(1.3)  & 2.3(0.8)  & 1.3(1.6)  & 49.8(2.8) & 49.2(1.4) & 50.9(1.7) & 47.0(2.2) & 46.4(2.8) \\
                              & Yelp                     & 6.3(0.3)  & 6.0(0.4)  & 6.1(0.2)  & 6.3(0.4)  & 6.0(0.3)  & 6.0(1.1)  & 5.2(1.1)  & 6.1(1.8)  & 6.8(0.9)  & 6.1(0.9)  & 60.9(1.1) & 59.2(1.8) & 59.9(0.6) & 60.6(2.2) & 58.8(2.0) \\
                              & RCT                      & 10.5(0.2) & 10.6(0.2) & 10.4(0.5) & 10.1(0.7) & 10.2(0.6) & 13.4(1.3) & 12.4(1.6) & 11.4(1.0) & 11.8(0.9) & 11.8(1.2) & 70.5(0.7) & 71.2(1.3) & 70.5(0.4) & 69.8(1.5) & 69.8(1.6) \\ \midrule
\multirow{3}{*}{LOF}          & AG                       & {\bf 9.3(0.0) }  & {\bf 8.1(0.0) }  & {\bf 7.5(0.0) }  & {\bf 6.7(0.0) }  & 5.8(0.0)   & {\bf 10.5(0.0) } & {\bf 8.4(0.0) }  & {\bf 7.4(0.0) }  & {\bf 7.4(0.0) }  & 3.2(0.0)   & {\bf 70.8(0.0) } & {\bf 68.5(0.0) } & {\bf 66.6(0.0) } & {\bf 63.3(0.0) } & 58.6(0.0)  \\
                              & Yelp                     & {\bf 7.2(0.0) }  & {\bf 7.0(0.0) }  & {\bf 6.9(0.0) }  & 6.8(0.0)   & 6.5(0.0)   & 7.9(0.0)   & 7.3(0.0)   & 7.3(0.0)   & 7.3(0.0)   & 6.7(0.0)   & 61.5(0.0)  & 61.0(0.0)  & 60.6(0.0)  & 59.9(0.0)  & 58.6(0.0)  \\
                              & RCT                      & {\bf 15.3(0.0) } & 12.9(0.0)  & 10.6(0.0)  & 9.0(0.0)   & 7.7(0.0)   & {\bf 24.6(0.0) } & {\bf 22.8(0.0) } & {\bf 18.0(0.0) } & 15.8(0.0)  & 14.0(0.0)  & {\bf 74.3(0.0) } & 72.7(0.0)  & 69.5(0.0)  & 63.8(0.0)  & 59.7(0.0)  \\ \midrule
\multirow{3}{*}{BERT-rank} & AG                       & 4.4(0.0)  & 4.3(0.0)  & 4.3(0.0)  & 4.2(0.0)    & 4.1(0.0)  & 2.1(0.0)  & 1.1(0.0)  &  1.1(0.0)  & 1.1(0.0)  & 1.1(0.0)  & 49.6(0.0)     & 49.2(0.0)  & 48.9(0.0)  & 48.3(0.0)  & 47.1(0.0)  \\
                              & Yelp                     &6.1(0.0)  &6.0(0.0)  &6.0(0.0)  &6.0(0.0)  & 5.9(0.0)   & 5.5(0.0) & 5.5(0.0)  &5.5(0.0) &5.5(0.0)  &5.5(0.0)  & 60.4(0.0)   & 60.3(0.0)  &60.1(0.0)  & 59.9(0.0)  & 59.3(0.0)    \\
                              & RCT                      &10.8(0.0)       &10.8(0.0)       &10.7(0.0)  & 10.7(0.0)  & 10.6(0.0)  & 11.4(0.0)  & 11.4(0.0)  &11.4(0.0)  & 11.4(0.0)  & 11.4(0.0)  & 71.4(0.0)   & 71.3(0.0)  &71.2(0.0)  & 71.1(0.0)  & 70.8(0.0)  \\ \midrule
\multirow{3}{*}{OC-SVDD}           & AG &6.8(0.3) &6.5(0.2) &5.8(0.2) &5.8(0.3)                                  &5.3(0.2) &7.8(2.0) &8.2(2.0) &6.5(0.8) &5.9(2.0)                               &4.6(1.4) &61.2(1.2) &60.0(0.4) &57.8(1.4) &57.1(1.7)                             &54.4(1.3)      \\
                              & Yelp &7.0(0.5) &6.9(0.2) &6.7(0.2) &{\bf 6.8(0.2)} &6.6(0.3) &{\bf 8.5(0.7)} &{\bf 8.5(0.9)} &7.6(1.1) &8.0(1.4) &7.2(0.8) &{\bf 61.8(1.1)} &{\bf 62.1(0.9)} &61.8(1.0) &61.7(0.9) &61.1(0.9)      \\
                              & RCT &13.3(0.4) &12.5(0.4) &11.8(0.4) &11.0(0.4) &10.6(0.4) &20.6(0.7) &19.6(1.2) &{\bf 18.5(1.1)} &16.7(0.7) &15.7(0.6) &72.5(0.5) &72.1(0.7) &71.0(1.1) &69.8(0.6) &68.7(0.7)\\ \midrule
\multirow{3}{*}{AI-SVDD}           & AG &5.5(0.4) &5.5(0.3) &5.6(0.2) &6.3(0.4)                                 &{\bf 6.8(0.1)} &2.9(1.4) &3.8(1.3) &4.0(0.8) &5.7(1.6)                                 &{\bf 6.1(1.2)} &56.9(2.2) &57.0(1.1) &57.4(1.1)                                        &60.3(1.4) &{\bf 62.7(0.6)}\\
                              & Yelp  &6.3(0.5) &6.5(0.3) &6.6(0.5) &{\bf 7.0(0.6)} &{\bf 7.6(1.4)} &6.5(1.2) &6.5(1.0) &{\bf 8.4(1.2)} &{\bf 8.6(1.1)} &{\bf 9.9(3.0)} &60.3(2.1) & 61.8(1.5) &{\bf 62.0(2.5)} &{\bf 62.1(2.1)} &{\bf 62.3(4.4)}\\
                              & RCT &12.3(0.8) &{\bf 13.2(0.4)} &{\bf 12.9(0.7)} &{\bf 13.3(0.8)} &{\bf 13.4(0.4)} &16.6(0.6) &18.2(2.0) &16.8(1.2) &{\bf 17.8(1.8)} &{\bf 18.9(0.8)} &73.4(1.3) &{\bf 74.7(0.5)} &{\bf 74.0(1.1)} &{\bf 75.6(1.0)} &{\bf 74.4(0.9)}\\ \bottomrule
\end{tabular}
\caption{Performance of 5 runs (mean and std.) on $5\%$ polluted test data with various polluted training data. Since there is no randomness in OC-SVM, LOF and BERT-rank, the standard deviations of these methods are zero.}
\label{tb:exp5.2-perf}
% \vspace{-0.3cm}
\end{table*}

\subsubsection{Dataset}
To evaluate the ability of detecting unknown classes, we conduct experiments on three datasets, which cover different domains: news (News Topic Categorization uses AG's news corpus~\cite{zhang2015character}), online reviews (Review Categorization uses Yelp dataset~\cite{zhang2015character}), and biomedical papers (Abstract Role Categorization uses RCT dataset~\cite{dernoncourt2017pubmed}). We create text anomaly datasets~\cite{larson2019outlier,ruff2019self} in the fashion of one-class classification setup. For each dataset we study, one class is used as normal and all the other classes are considered as anomalies. Examples with the normal class is relabelled as $y=1$ ("normal") and examples from remaining classes are relabelled as $y=-1$ ("anomaly"). Detailed descriptions of how each domain data for unknown class detection task is generated are as follows:
%in appendix~\ref{app:data2}.
{\bf News Topic Categorization} is a topic classification task. We use the AG's news corpus~\cite{zhang2015character}, which groups the news into 4 categorizations: 'world', 'sports', 'business' and 'sci/tech'. 'sports' topic is chosen as the normal, other topics are used as anomaly. Since the news contains both headlines and content, we consider using the headlines to detect text anomalies. Removing invalid headline text results in $28749$ normal training samples, $1250$ normal development samples and $1900$ normal test samples.
% In other words, we want to identify world, business or sci/tech news from a set of sports news.
{\bf Review Categorization} is a task to predict the stars the user has given based on their text reviews. We use Yelp dataset~\cite{zhang2015character}. There are five labels in total, from 1 to 5 stars. The 5 star is chosen as the normal and 1-4 stars are used as anomaly. Since the dataset is large (training set contains $650k$ samples), we filter out the text samples that are over 50 words or less than 3 words and use the filtered as the input text. The resulting number of normal training samples is $31856$, normal development sample size is $11622$, and normal test sample size is $3295$.
% We want to identify negative and normal reviews from the strong positive reviews.
{\bf Abstract Role Categorization} is a task to predict the role of a text in abstracts. We use RCT dataset~\cite{dernoncourt2017pubmed}, in which the role is labelled with five classes: 'background', 'objective', 'method', 'results' and 'conclusions'. The 'conclusions' role is considered as normal, and the others are considered as anomaly. We use the raw text to distinguish anomaly from the normal. The normal training sample size is $27168$, normal development sample size is $4582$, normal test sample size is $4571$.

% \vspace{-0.1cm}
\subsubsection{Setting}
To mimic the real situation where only a small number of anomaly text exists, we follow a similar setting of the experiments in Section~\ref{subsec:5.1setting}. 
Specifically, we adopt the same range of pollution proportion $p$ to generate training datasets and the same $5\%$ pollution setting is used for the development and the test datasets. Anomaly samples are selected uniformly from all classes other than the one normal class. 
% Table~\ref{tb:cls-datasets} summarizes statistics of the datasets after pre-processing.
% In this experiment, we train the models in different $p$ portion data and evaluate on $5\%$ anomaly proportion data using MAP, Recall@5 and AUC. 

% \paragraph{Baseline I: Unbalanced binary classification} we consider this anomaly detection as a unbalanced binary classification problem, in which we  train and evaluate using normal classification scheme. we use SVM~\cite{chang2011libsvm}, multiple layer perceptron (MLP) and BERT~\cite{devlin2019bert} with fine-tuning as baseline classifiers. SVM and MLP are trained with scikit-learn library~\cite{scikit-learn}. We obtain BERT model using huggingface library~\cite{Wolf2019HuggingFacesTS}. Precision, recall and AUC are used a evaluation metrics.

% \paragraph{Baseline II: One-class classification}
% An alternative setting is a anomaly detection framework, in which we predict the anomalous using one-class classifiers.
% One-class SVM~\cite{li2003improving} is used as the baseline.  In this setting, Mean average precision(MAP), recall@k  and area under receiver operating characteristic (AUC) are used as evaluation metrics. 

% \vspace{-0.1cm}
\subsubsection{Result and analysis}

Table~\ref{tb:exp5.2-perf} summarizes the anomaly detection performance in terms of MAP, Recall@5 and AUC on the $5\%$-polluted test data on different approaches. 
We can observe from the table that all of the unsupervised methods (OC-SVM, ISF, LOF, BERT-rank, OC-SVDD) exhibit a performance decreasing when the training pollution increase, 
while our proposed deep AI-SVDD approach demonstrates increased performance when pollution proportion is increasing. 
For the AG dataset, the LOF method with BERT embedding achieves the highest performance of all metric when pollution is less than $8\%$. Our deep AI-SVDD approach competes all the others when the training pollution is $8\%$. 
% We believe by increasing the training pollution, we can still increase the performance more. 
For the Yelp dataset, when pollution is less than $4\%$, the highest values for different metrics are sparsely located for various unsupervised methods, but our AI-SVDD approach shows consistent highest performance on all metric when training pollution $p\geq 4\%$. 
% For the RCT dataset, even LOF has the highest value for Recall@5 when $p\leq 4\%$, our proposed method competes all methods on MAP and AUC when $p>0\%$. 
For the RCT dataset, our AI-SVDD method outperforms all other models in terms of MAP and AUC when $p>0\%$, and show competitive performance in terms of Recall@5 when compared with LOF.

% \vspace{-0.2cm}
\subsection{Medical document anomaly detection}
% \vspace{-0.1cm}
\subsubsection{Dataset}

In the last, we evaluate our approach on a real-world text data quality control task. Specifically, our target is to identify qualified question answering (QA) pairs from a set of multi-source multi-form medical text data in Chinese. Disqualified samples include low-quality QA pairs (such that the question is incomplete or it does not exhibit a clear intention), or non-QA data such as scientific essays, {\it etc}. 
% Please see table~\ref{tb:med-dataset} for an example.
In this task, we just focus on the title or question part of the data. The original crowdsourcing data contains over 10 million samples that need to be examined, but only a small portion of the data is labeled. As a result, such labeled multi-source data contains a total of $7,836$ normal samples and $384$ anomaly samples from $13$ data sources. 
% \begin{table}[ht]
% \small
% \centering
% \begin{tabular}{@{}cl@{}}
% \toprule
% Label & Example  \\ \midrule
%  1    & \language[cn]{肝硬化是什么}  \\
%  1    & \language[cn]{阿莫西林的功效与主治}  \\
%  1    & \language[cn]{我的牛皮癣治好了} \\
%  -1    & \language[cn]{马应龙痔疮膏}       \\ 
%  -1    & \language[cn]{睡前一个动作暴瘦肚子}\\
%  -1    & \language[cn]{屁股骨质增生痛}\\\bottomrule
% \end{tabular}
% \caption{Examples of the Medical data with $1$ for normal class and $-1$ for anomaly.}\label{tb:med-dataset}
% \end{table}

% \vspace{-0.1cm}
\subsubsection{Setting}
To demonstrate the performance of the proposed approach and the competing methods on various test datasets, we use $5$ different random splits of the total $8,220$ data samples. To maintain a similar class ratio on training and test data, we combine $70\%$ of the normal samples and $70\%$ anomaly samples as the training data, and the remaining $30\%$ of both normal and anomaly samples are considered as the test data. For each of the $5$ random splits, we denote it as one Monte-Carlo (MC) run. %, we train all models on the training data and evaluate the performance on the test data.

% \vspace{-0.1cm}
\subsubsection{Result and analysis}
Table~\ref{tb:exp5.3-perf} shows the mean and standard deviation of the anomaly detection performance on various methods in terms of MAP, Recall@5 and AUC of $5$ independent MC runs. % of the test data. 
We can observe from the table that traditional outlier detection approaches (OC-SVM, ISF, LOF) even with BERT embeddings, do not show promising results on any metrics, especially for the OC-SVM. 
The BERT-rank method can compete with the traditional approaches and show increased performance. 
In addition, the deep OC-SVDD method has a $1\%$ increase in terms of MAP and AUC and a $5\%$ increase in Recall@5 compared with the BERT-rank. 
Moreover, our proposed AI-SVDD approach outperforms all comparing methods in all metric in this medical text anomaly detection task. Particularly, our method improves the average MAP by $16\%$, the average Recall@5 by $14\%$ and the average AUC by $2.4\%$, when compared with the OC-SVDD approach.

\begin{table}[ht]
\setlength{\tabcolsep}{2pt}
\fontsize{7}{9}\selectfont
\centering
\begin{tabular}{@{}cccc|cccc@{}}
\toprule
Method & MAP   & Recall@5    & AUC & Method & MAP   & Recall@5    & AUC   \\\midrule
OC-SVM& 7.1(0.9) & 11.0(2.0)   & 54.2(1.2) & BERT-rank& 19.1(1.6) & 23.1(1.1)   & 78.9(1.6) \\
ISF& 15.1(1.9) & 21.7(3.4)   & 77.4(1.7) & OC-SVDD & 20.0(1.7) & 28.3(2.7) & 80.1(1.3)\\
LOF& 13.6(1.8) & 16.7(2.3)   & 76.6(2.9) & AI-SVDD & {\bf 35.9(3.5)} &  {\bf 42.2(4.5)}   & {\bf 82.5(2.2)}\\\bottomrule
\end{tabular}
\caption{Medical document title anomaly detection performance of 5 MC runs.}\label{tb:exp5.3-perf}
% \vspace{-0.5cm}
\end{table}

\subsection{Discussion and future work}
We want to emphasize that our approach is not a pure supervised classification method but a variant of the deep SVDD approach (which is pure unsupervised). The main idea is when a small number of negative examples (which belong to anomaly and should be rejected) are available, they can be incorporated into the training to improve the data description. Although the objective uses a small amount of the anomaly label y, the target and our objective still focuses on one class and fit a good data description to capture its distribution. In the proposed experiments, We only compare our approach with traditional outlier detection approaches and several other deep SVDD approaches. Due to space limitation and slight beyond our concentration of this work, we did not show the comparison of our approach with the classification approach, those will remain as the future work. We believe that the proposed method can beat the performance of the classification approach, especially under the scenario that the majority of the test negative samples are not observed and not from the sample distribution as in the training. Please see Figure 1 for an illustrative example. This is due to the fact that the negative samples do not contain a consistent pattern and come from the same distribution, it is hard to model both positive and negative samples to train a classifier.

\section{Conclusion}
\label{sec:conc}
In this work, we developed a novel deep AI-SVDD model that aims to discriminate the normal class data from anomalies or outliers. 
% We carefully examined the deep OC-SVDD objective and proposed a revision with a corresponding detailed analysis in the case study.
We carefully examined the deep OC-SVDD objective and proposed a revision with a center updating mechanism.
To tackle the text anomaly detection task, we proposed a systematic deep model with a solution for constraints minimization. 
We employed several baseline methods and developed a competing deep OC-SVDD model from scratch.  
To evaluate our proposed approach, we conducted our experiments in three different applications with textual datasets. 
Experimental results demonstrated that our proposed AI-SVDD approach is promising and it could provide competitive results when data is polluted. AI-SVDD outperforms various comparing approaches on the three proposed applications.

% The results demonstrated that the proposed AI-SVDD approach, even though did not show promising results compared with deep OC-SVDD when data is not polluted, show competing results when data is polluted and outperforms various comparing approaches for the three proposed applications.

\section*{Acknowledgments}
We would like to show our special thanks to Xingyuan Pan, who is a Ph.D. graduated from University of Utah and now working as an applied scientist in Amazon. He provides meaningful suggestions and tremendous help for this work. \\

% % Entries for the entire Anthology, followed by custom entries
% \bibliography{aaai22}
% % \bibliographystyle{aaai22.bst}

\appendix

\section{Appendix}
\label{sec:appendix}

For simplicity, we ignore the ranges information in sum expressions in the discussions below, i.e. for a sample index $h\in \{i, j, k\}$, $\sum_{h}$ stands for $\sum_{h=1}^n$, \& for a layer index $l$, $\sum_{l}$ stands for $\sum_{l=1}^L$.

% \subsection{Proof of equality between Eqs.~\eqref{eq:oc-joint} \& \eqref{eq:oc-joint-eq}}
\subsection{Proof of equality between Eqs. (2) \& (3) in main paper}
\label{app:simplify1}

Since ${\bf c}^*=\frac{1}{n}\sum_{j} \phi({\bf x}_j ; {\cal W})$ in % Eq.~\eqref{eq:oc-joint}, 
Eq. (2) in main paper,
after plugging it back into % Eq.~\eqref{eq:oc-joint}, 
Eq. (2) in main paper,
we can obtain:
\begin{align*}
\min_{{\cal W}} L({\cal W}) = \tfrac{1}{n} \sum\nolimits_{i}& \|\phi({\bf x}_i; {\cal W})-\tfrac{1}{n}\sum\nolimits_{j} \phi({\bf x}_j ; {\cal W})\|_2^2 \\
& + \tfrac{\lambda}{2}\sum\nolimits_{l}\|{\bf W}^l\|_F^2.
\end{align*}
Let ${\bf u}_i = \phi({\bf x}_i ; {\cal W}), {\bf u}_j = \phi({\bf x}_j ; {\cal W})$, and ${\cal U} =\{{\bf u}_1, \ldots, {\bf u}_n\}$. Denote $f({\cal U})$ be the first term of $L({\cal W})$, then it can be simplified as:
% and $f({\cal U}) = \frac{1}{n}\sum_{i}\|{\bf u}_i-\frac{1}{n}\sum_{j}{\bf u}_j\|_2^2$.
\begin{align*}
f({\cal U}) = & \tfrac{1}{n}\sum\nolimits_{i}\|{\bf u}_i - \tfrac{1}{n}\sum\nolimits_{j}{\bf u}_j\|_2^2\\
            = & \tfrac{1}{n}\bigl( \sum\nolimits_{i} \|{\bf u}_i\|_2^2- 2\sum\nolimits_{i} {\bf u}_i^T (\frac{1}{n} \sum\nolimits_{j} {\bf u}_j)\\
              & +\tfrac{1}{n} \|\sum\nolimits_{j} {\bf u}_j\|^2\bigr)\\
            = & \tfrac{1}{n}\sum\nolimits_{i} \|{\bf u}_i\|_2^2- \tfrac{2}{n^2}(\sum\nolimits_{i} {\bf u}_i)^T (\sum\nolimits_{j} {\bf u}_j)\\
              & + \tfrac{1}{n^2}\|\sum\nolimits_{j} {\bf u}_j\|^2_2\\
            = & \tfrac{1}{n}\sum\nolimits_{i} \|{\bf u}_i\|_2^2-\tfrac{1}{n^2}(\sum\nolimits_{i} {\bf u}_i)^T (\sum\nolimits_{j} {\bf u}_j). \numberthis \label{eq:simplify}
\end{align*}
Substituting \[\sum\nolimits_{i} \|{\bf u}_i\|_2^2 = \tfrac{1}{2n}(\sum\nolimits_{i} \sum\nolimits_{j}\|{\bf u}_i\|_2^2+\sum\nolimits_{i} \sum\nolimits_{j}\|{\bf u}_j\|_2^2)\] into the first term of (\ref{eq:simplify}), yields:
\begin{align*}
f({\cal U}) = & \frac{1}{2n^2}(\sum_{i,j} \|{\bf u}_i\|_2^2+\sum_{i,j}\|{\bf u}_j\|_2^2-2\sum_{i,j} {\bf u}^T_i{\bf u}_j)\\
            = & \frac{1}{2n^2}\sum\nolimits_{i,j} \| {\bf u}_i- {\bf u}_j\|_2^2.
\end{align*}
Therefore, with ${\bf c}^*$, minimizing % Eq.~\eqref{eq:oc-joint} 
Eq. (2)
is equivalent to minimizing % Eq.~\eqref{eq:oc-joint-eq}.
Eq. (3) in main paper.

% \subsection{Proof of equality between Eqs.~\eqref{eq:bc-joint} \& \eqref{eq:bc-joint-eq}}
\subsection{Proof of equality between Eqs. (5) \& (6) in main paper}
\label{app:simplify2}

Since ${\bf c}^*= \frac{\sum_{i} y_i \phi({\bf x}_i ; {\cal W})}{\sum_{i} y_i}$ in % Eq.~\eqref{eq:bc-joint}, 
Eq. (5) in main paper, 
after plugging it back into % Eq.~\eqref{eq:bc-joint-eq}, 
Eq. (6) in main paper, 
we can obtain:
\begin{align*}
\min_{{\cal W}} & \frac{1}{n} \sum_{i} y_i \|\phi({\bf x}_i ; {\cal W})-\frac{\sum_{j} y_j \phi({\bf x}_j ; {\cal W})}{\sum_{j} y_j}\|_2^2,\\
& \text{s.t.~}\|{\bf W}^l\|_F^2, ~\forall l=1,2,\ldots, L.
\end{align*}
If we denote ${\mathbold \phi}_i = \phi({\bf x}_i ; {\cal W}), {\mathbold \phi}_j = \phi({\bf x}_j ; {\cal W})$, and
$L({\cal W})=\frac{1}{n}\sum_{i} y_i \|\phi({\bf x}_i ; {\cal W})-\frac{\sum_{j} y_j \phi({\bf x}_j ; {\cal W})}{\sum_{j} y_j}\|_2^2$, we can simplify it as follows:
\begin{align}\label{eq:simplify2}
\nonumber L({\cal W}) =& \frac{1}{n}\bigl(\sum_{i} y_i\|{\mathbold \phi}_i\|_2^2 - 2\frac{\sum_{i} y_i {\mathbold \phi}_i^T (\sum_{j} y_j {\mathbold \phi}_j)}{\sum_{j} y_j} \\
\nonumber &+ \frac{\|\sum_{j} y_j {\mathbold \phi}_j\|^2}{\sum_{j} y_j} \bigr)\\
=&\frac{1}{n}\bigl(\sum_{i} y_i\|{\mathbold \phi}_i\|_2^2 - \frac{(\sum_{i} y_i {\mathbold \phi}_i)^T (\sum_j y_j {\mathbold \phi}_j)}{\sum_j y_j}\bigr).\hspace{0.3cm}
\end{align}
Since $\sum_j y_j=\sum_k y_k$, and \[\sum_{i} y_i \|{\mathbold \phi}_i\|_2^2=\frac{1}{2}\frac{\sum_{i,j} y_i y_j\|{\mathbold \phi}_i\|_2^2 +\sum_{j,i} y_j y_i \|{\mathbold \phi}_j\|_2^2}{\sum_{j} y_j},\]
substituting it into the first term of (\ref{eq:simplify2}), yields
\begin{align*}
\nonumber L({\cal W}) =&
\frac{1}{2n\sum_{k} y_k}(\sum_{i,j} y_i y_j\|{\mathbold \phi}_i\|_2^2+\sum_{i,j} y_i y_j\|{\mathbold \phi}_j\|_2^2\\
& - 2\sum_{i,j} y_i y_j {\mathbold \phi}^T_i{\mathbold \phi}_j)\\
=&\frac{1}{2n\sum_{k} y_k}\sum_{i,j} y_i y_j\| {\mathbold \phi}_i- {\mathbold \phi}_j\|_2^2.
\end{align*}
Therefore, with ${\bf c}^*$, minimizing % Eq.~\eqref{eq:bc-joint} 
Eq. (5) is equivalent to minimizing % Eq.~\eqref{eq:bc-joint-eq}.
Eq. (6) in main paper.

\subsection{Parameter tuning}
\label{app:paratune}
We tune the hyper-parameters for the deep OC-SVDD network by setting hidden size $hs \in \{128, 256, 512, 1024\}$, latent size $ls \in \{32, 64, 128, 256\}$, batch size $bs \in \{32, 64, 128, 256\}$ and regularization term $\lambda \in \{0, 1e-4, 1e-2, 1, 10\}$. For our proposed AI-SVDD network, we search over a larger range for each hyper-parameter, hidden size $hs \in \{128, 256, 512,1024,2048,4096\}$, latent size $ls \in \{16, 32, 64, 128, 256, 512\}$ and batch size $bs \in \{16, 32, 64, 128, 256, 512\}$. 
Following the cross-validation mechanism, we find the optimal hyper-parameters for each application using the non-polluted training data and the corresponding development data for validation. For all three applications, the learning rate is fixed as $lr=0.001$ and the epoch is fixed to be $3$. For the text topic change application, the optimal setting for deep OC-SVDD network is $hs=256,ls=128, bs=64,\lambda=1e-4$ and the optimal setting for our deep AI-SVDD network is $hs=2048,ls=256, bs=128$. For the text anomaly detection application, the optimal setting for deep OC-SVDD network in AG dataset is $hs=512,ls=64, bs=256,\lambda=1e-4$, Yelp and RCT-20k dataset are both $hs=512,ls=64, bs=32,\lambda=1e-4$. The optimal setting for our deep AI-SVDD network with AG dataset is $hs=1024,ls=128, bs=256$, with Yelp dataset is $hs=512,ls=64, bs=64$, and with RCT-20k dataset is $hs=512,ls=256, bs=128$.
For the application of medical document title anomaly detection, the optimal setting for deep OC-SVDD network is $hs=512,ls=128, bs=64,\lambda=1e-4$ and the optimal setting for our deep AI-SVDD network is $hs=2048,ls=64, bs=64$.

\section{Supplemental Material}
\label{sec:supplemental}

\subsection{Case study--I: Vanilla network}
\label{supp:case_study}
For preliminary exploration purpose, we consider a very simple neural network with a linear layer. Let ${\bf W}_1$ be the weight to be learned in % Eq.~\eqref{eq:oc-svdd}
Eq. (1) in main paper 
and $\phi({\bf x}_i ; {\cal W})={\bf W}^T{\bf x}_i$, then we have:
\begin{equation}\label{eq:oc-svdd-simp}
    \min_{{\bf W}_1} \frac{1}{n}\sum_{i=1}^n\|{\bf W}_1^T{\bf x}_i-{\bf c}\|_2^2 + \frac{\lambda}{2}\|{\bf W}_1\|_F^2.
\end{equation}
An optimal solution of Eq.~\eqref{eq:oc-svdd-simp} can be obtained by ${\bf W}_1^*=(\frac{2}{n}\sum_{i=1}^n{\bf x}_i{\bf x}^T_i + \lambda {\bf I}_D)^{-1}(\frac{2}{n}\sum_{i=1}^n{\bf x}_i{\bf c}^T)$.

Similarly, for % Eq.~\eqref{eq:oc-joint-eq}, 
Eq. (3) in main paper, 
let ${\bf W}_2$ be the weight to be learned and $\phi({\bf x}_i ; {\cal W})={\bf W}_2^T{\bf x}_i$, then problem %\eqref{eq:oc-joint-eq} 
(3) in main paper 
is equivalent to:
\begin{equation}\label{eq:oc-svdd-c-simp}
\min_{{\bf W}_2} \frac{1}{2n^2}\sum_{i,j=1}^n \| {\bf W}_2^T ({\bf x}_i- {\bf x}_j)\|_2^2+ \frac{\lambda}{2}\|{\bf W}_2\|_F^2.
\end{equation}
Taking the derivative of Eq.~\eqref{eq:oc-svdd-c-simp} 
{\it w.r.t.} ${\bf W}_2$ and assigning the gradient to zero, we have:
\begin{equation*}
(\frac{1}{n^2}\sum_{i,j=1}^n({\bf x}_i- {\bf x}_j)({\bf x}_i- {\bf x}_j)^T + \lambda {\bf I}) {\bf W}_2 = {\bf 0}.
\end{equation*}
Let $A=\frac{1}{n^2}\sum_{i,j=1}^n({\bf x}_i- {\bf x}_j)({\bf x}_i- {\bf x}_j)^T + \lambda {\bf I}$, then an optimal solution of Eq.~\eqref{eq:oc-svdd-c-simp}, ${\bf W}_2^*=[{\bf w}_1, \ldots, {\bf w}_d]$, falls into the null space of $A$. However, since $A$ is full-rank, the optimal ${\bf W}_2^*$ in such a case can only be zero. In other words, the network learns nothing but projects all data points into one.

To avoid the aforementioned problem, we consider replacing the regularization term in Eq.~\eqref{eq:oc-svdd-c-simp} with a constraint $\|{\bf W}_2\|^2_F = 1$ instead, then the problem can be newly defined as:
\begin{equation}\label{eq:oc-joint-new}
\min_{{\bf W}_2} \frac{1}{2n^2}\sum_{i,j=1}^n \|{\bf W}_2^T ({\bf x}_i- {\bf x}_j)\|_2^2,\text{~s.t.~}\|{\bf W}_2\|^2_F=1.
\end{equation}
Since $\| {\bf W}_2^T ({\bf x}_i- {\bf x}_j)\|_2^2 \leq \| {\bf W}_2\|_F^2\|{\bf x}_i- {\bf x}_j\|_2^2=\|{\bf x}_i- {\bf x}_j\|_2^2$, by applying the Cauchy-Schwarz inequality in~\citet{bhatia1995cauchy}, we can easily show that the transformed feature points (${\bf W}_2^T{\bf x}_i$) would have a more compact enclosure than the original data points. 

\subsection{Case study--II: Comparison of different objectives}
\label{supp:dis-com}
In this part, we aim to illustrate the difference of one-class objectives in Eq.~\eqref{eq:oc-svdd-simp} \& \eqref{eq:oc-joint-new} and compare with our proposed AI-SVDD objective. 
Followed by the vanilla network in Eqs.~\eqref{eq:oc-svdd-simp}, we also employ a simple one linear layer network  $\phi({\bf x}_i ; {\cal W})={\bf W}_3^T{\bf x}_i$, then the objective~%\eqref{eq:bc-joint-eq} 
(6)
becomes:
\begin{align}\label{eq:bc-joint-simp}
    \nonumber \min_{{\cal W}}  & \frac{1}{2n\sum_{k=1}^n y_k}\displaystyle{\sum_{i,j=1}^n} y_i y_j \| {\bf W}_3^T({\bf x}_i- {\bf x}_j)\|_2^2\\
   & \text{s.t.~} \|{\bf W}_3\|_F^2 = 1, ~\forall l=1,2,\ldots,L.
\end{align}

\begin{figure}[t]
\vspace{-2.5cm}
\begin{minipage}[b]{1\linewidth}
  \centering
  \centerline{\includegraphics[width=8cm]{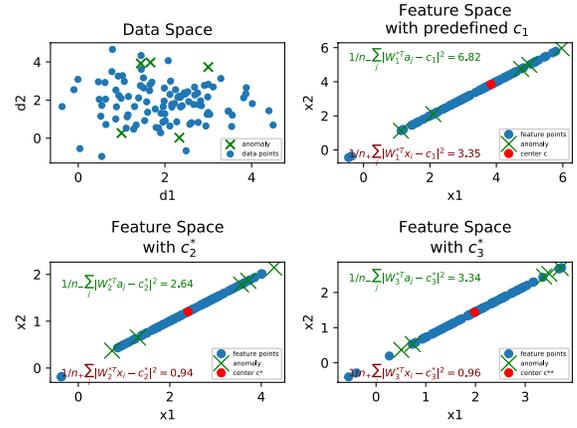}}
 \vspace{-2.5cm}
\end{minipage}
\caption{Illustrations on three different linear transformations: top-left is the original data points in input space; top-right is the transformed points of Eq.~\eqref{eq:oc-svdd-simp} in feature space; bottom-left is the transformed points of Eq.~\eqref{eq:oc-joint-new} in feature space; bottom-right is the transformed points of Eq.~\eqref{eq:bc-joint-simp} in feature space.}
\label{fig:trans-diff}
\end{figure}
To compare and illustrate the differences among objectives, we generate a small number $n_{+}=100$ data points from a Gaussian distribution where each point ${\bf x}_i \sim {\cal N}([2, 2]^T, {\bf I})$. 
Next, We obtain the optimal linear transformation $W_1^*$ for Eq.~\eqref{eq:oc-svdd-simp} as well as solve the constraint optimization in Eq.~\eqref{eq:oc-joint-new} with the optimals ${\bf W}_2^*$ and ${\bf c}_2^{*} = \frac{2}{n}\sum_{i=1}^n {\bf W}_2^{*T}{\bf x}$. Minimizing Eq.~\eqref{eq:bc-joint-simp} {\it w.r.t.} to ${\bf W}_3$, we also obtain the optimal transformation ${\bf W}_3^{*}$ and the corresponding optimal center ${\bf c}_3^{*}=\frac{\sum_{i=1}^n y_i {\bf W}_3^{*T}{\bf x}_i}{\sum_{i=1}^n y_i}$.
To simulate anomaly points, we randomly generated $n_{-}=5$ points that are sampled from a circle centered at $[2, 2]^T$ with radius $2$. For the sake of exploring the discriminative powers in a harder and more realistic situation, we carefully designed the anomaly points to be close to some of the edge points (located 2$\sigma$ away from the Gaussian center). 

In Figure~\ref{fig:trans-diff}, we illustrate the original data space in the top-left part of the figure and the blue dots refer to normal points and green crosses refer to the abnormal points. 
The linear transformed data points (${\bf W}_1^{*T}{\bf x}_i$s and ${\bf W}_1^{*T}{\bf a}_j$s where ${\bf a}_j$s refer to the $5$ anomaly points) are shown in the top-right of Figure~\ref{fig:trans-diff}, and the linear transformed data points (${\bf W}_2^{*T}{\bf x}_i$s and ${\bf W}_2^{*T}{\bf a}_j$s, where ${\bf W}_2^{*T}$ is obtained from minimizing Eq.~\eqref{eq:oc-joint-new}) are shown in the bottom-left of Figure~\ref{fig:trans-diff}. In Figure~\ref{fig:trans-diff}, the data hypersphere of Eq.~\eqref{eq:oc-joint-new} as $\frac{1}{n_\texttt{+}}\sum_{i} \|{\bf W}_2^{*T}{\bf x}_i-{\bf c}_2^{*}\|_2^2=0.94$ in the feature space is actually much smaller than the data hypersphere of Eq.~\eqref{eq:oc-svdd-simp} as $\frac{1}{n_\texttt{+}}\sum_{i} \|{\bf W}_1^{*T}{\bf x}_i-{\bf c}_1\|_2^2=3.35$. In addition, the ratio between the average squared distance (between the anomaly points and the center) and that of the normal points $\frac{\frac{1}{n_\texttt{-}}\sum_{j} \|{\bf W}_2^{*T}{\bf a}_j-{\bf c}_2^*\|_2^2}{\frac{1}{n_\texttt{+}}\sum_{i} \|{\bf W}_2^{*T}{\bf x}_i-{\bf c}_2^*\|_2^2}=\frac{2.64}{0.94}=2.8$ in Eq.~\eqref{eq:oc-joint-new} is larger than $\frac{\frac{1}{n_\texttt{-}}\sum_{j} \|{\bf W}_1^{*T}{\bf a}_j-{\bf c}_1\|_2^2}{\frac{1}{n_\texttt{+}}\sum_{i} \|{\bf W}_1^{*T}{\bf x}_i-{\bf c}_1\|_2^2}=\frac{6.82}{3.35}=2.0$ in Eq.~\eqref{eq:oc-svdd-simp}. 

Since the AI-SVDD objective contains the anomaly data in the training, we have to randomly generated $5$ additional anomaly points from the same anomaly points distribution and use them for model training. In Figure~\ref{fig:trans-diff}, we plot the transformed points in the bottom-right sub-figure. 
The data hypersphere of Eq.~\eqref{eq:bc-joint-simp} as $\frac{1}{n_\texttt{+}}\sum_{i} \|{\bf W}_3^{*T}{\bf x}_i-{\bf c}_3^{*}\|_2^2=0.96$ in the feature space is very close to the data hypersphere of Eq.~\eqref{eq:oc-joint-new}. In addition, the ratio between the average squared distance (between the anomaly points and the center) and that of the normal points $\frac{\frac{1}{n_\texttt{-}}\sum_{j} \|{\bf W}_3^{*T}{\bf a}_j-{\bf c}_3^*\|_2^2}{\frac{1}{n_\texttt{+}}\sum_{i} \|{\bf W}_3^{*T}{\bf x}_i-{\bf c}_3^*\|_2^2}=\frac{3.34}{0.96}=3.48$ in Eq.~\eqref{eq:bc-joint-simp} is even larger than $2.8$ in Eq.~\eqref{eq:oc-joint-new}. 

Hence, our proposed AI-SVDD model learns a transformation that introduces a more compact enclosure of the data hypersphere (compared to Eq.~\eqref{eq:oc-svdd-simp}), and in the meanwhile, discriminates the anomalies from the normal with a larger distance ratio (compared to Eq.~\eqref{eq:oc-joint-new}). 
Since the formulation in %Eq.~\eqref{eq:bc-joint} 
Eq. (5) of the main paper naturally introduces a larger distance between the anomaly and the center than that between the normal and the center, the difference (distance) between anomaly and the normal would be larger.

\subsection{Illustration on center difference}
\label{supp:ill-centerdiff}

\begin{figure}[t]
\vspace{-2cm}
\begin{minipage}[b]{1\linewidth}
  \centering
  \centerline{\includegraphics[width=7cm]{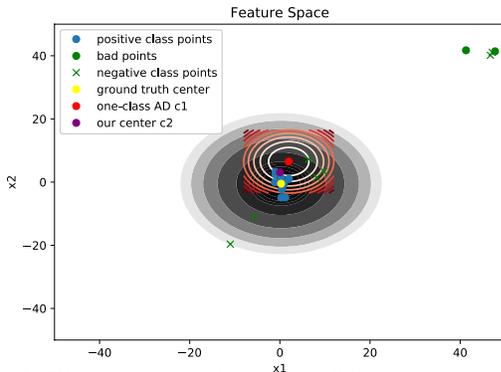}}
 \vspace{-2.5cm}
\end{minipage}
\caption{Illustration on the center difference between %Eq.~\eqref{eq:bc-joint} and Eq.~\eqref{eq:oc-joint-alt}}
Eq. (5) and Eq.(4) in main paper.}
\label{fig:center-diff}
\end{figure}
To illustrate the center difference between % Eq.~\eqref{eq:bc-joint} and Eq.~\eqref{eq:oc-joint-alt} 
Eq. (5) and Eq. (4) in main paper, 
and show how the label affecting the center, we randomly generate $10$ Gaussian points (normal class points) from ${\cal N}([0, 0]^T, 5{\bf I})$, and randomly sampled $2$ bad data from ${\cal N}([45, 45]^T, 10{\bf I})$, which are far from the normal class points. In the meantime, we also random generated $7$ anomaly class points, $2$ points are generated from ${\cal N}([45, 45]^T, 10{\bf I})$, $3$ points from ${\cal N}([10, 5]^T, 3{\bf I})$, $1$ point from ${\cal N}([-5, -10]^T, 2{\bf I})$ and $1$ point from ${\cal N}([-10, -20]^T, 2{\bf I})$. Please see the blue dot points for the normal class, green dot for the normal class bad data and green cross for the anomaly class in Figure~\ref{fig:center-diff}. 

The one class anomaly detection center $c_1$ in %Eq.~\eqref{eq:oc-joint-alt} 
Eq. (4)
is the pure average of all feature points, which is shown as the red dot. The binary class case center $c_2$ in %Eq.~\eqref{eq:bc-joint} 
Eq. (5)
is not the pure average, which is shown as the purple dot, where the yellow dot is the ground truth normal class center.
From Figure~\ref{fig:center-diff}, we can observe that the one class center is far from the ground truth, which means it is sensitive to the bad points, where the bad points may naturally comes from the data since the data itself might already contain small number of anomalies or outliers. 
In the meanwhile, if we use the one class center to perform a detection, the performance would be low since it can not distinguish some of the anomalies from the normal class. See Figure~\ref{fig:center-diff} for an example. 
However, since the binary class center ${\bf c}^* = \frac{\sum_{i=1}^n y_i \phi({\bf x}_i ; {\cal W})}{\sum_{i=1}^n y_i}$ is the difference between all of the normal class points and all of the anomaly points, the points close to each other with different labels would not contribute to the final center.
Therefore, the center in the binary class objective %Eq.~\eqref{eq:bc-joint} 
Eq. (5) 
is more robust to outliers of the data than the one class in %Eq.~\eqref{eq:oc-joint-alt}.
Eq. (4) in main paper.

\bibliography{aaai22}

\end{document}